\documentclass[10pt,twocolumn,letterpaper]{article}

\usepackage{iccv}
\usepackage{times}
\usepackage{epsfig}
\usepackage{graphicx}
\usepackage{amsmath}
\usepackage{amssymb}
\usepackage{lib}
\usepackage{caption}
\usepackage{subcaption}

\usepackage{lipsum}
\usepackage{epigraph}
\usepackage{multirow}
\usepackage{booktabs}
\usepackage{lib}
\usepackage{tabularx}
\usepackage{colortbl}
\usepackage{epstopdf}
\usepackage{flushend}
\usepackage{tabularx}

\newcommand{\vertical}[1]{\rotatebox[origin=l]{90}{\parbox{2.5cm}{#1}}}
\newcommand{\instr}[0]{{instr}}
\newcommand{\object}[0]{{obj}}

\newcommand{\vgg}[0]{{\small VGG}\xspace}
\newcommand{\rcnn}[0]{{\small R-CNN}\xspace}
\newcommand{\fastrcnn}[0]{{\small Fast R-CNN}\xspace}
\newcommand{\cnn}[0]{{\small CNN}\xspace}
\newcommand{\vcoco}[0]{{\small V-COCO}\xspace}
\newcommand{\vb}[1]{{\small \texttt{#1}}\xspace}
\newcommand{\vsrl}[0]{{\small VSRL}\xspace}
\newcommand{\train}[0]{\textit{train}\xspace}
\newcommand{\test}[0]{\textit{test}\xspace}
\newcommand{\val}[0]{\textit{val}\xspace}
\newcommand{\coco}[0]{{\small COCO}\xspace}
\newcommand{\amt}[0]{{\small AMT}\xspace}
\newcommand{\insertA}[2]{\IfFileExists{#2}{\includegraphics[width=#1\textwidth]{#2}}{\includegraphics[width=#1\textwidth]{figures/blank.jpg}}}
\newcommand{\insertB}[2]{\IfFileExists{#2}{\includegraphics[height=#1\textwidth]{#2}}{\includegraphics[height=#1\textwidth]{figures/blank.jpg}}}

\usepackage[pagebackref=true,breaklinks=true,letterpaper=true,colorlinks,bookmarks=false]{hyperref}

\iccvfinalcopy

\begin{document}

\title{Visual Semantic Role Labeling}

\author{Saurabh Gupta \\ UC Berkeley \\ {\tt\small sgupta@eecs.berkeley.edu} 
\and Jitendra Malik \\ UC Berkeley \\ {\tt\small malik@eecs.berkeley.edu} }

\maketitle

\begin{abstract} In this paper we introduce the problem of Visual Semantic Role
Labeling: given an image we want to detect people doing actions and localize
the objects of interaction. Classical approaches to action recognition either
study the task of action classification at the image or video clip level or at
best produce a bounding box around the person doing the action. We believe such
an output is inadequate and a complete understanding can only come when we are
able to associate objects in the scene to the different semantic roles of the
action. To enable progress towards this goal, we annotate a dataset of 16K
people instances in 10K images with actions they are doing and associate
objects in the scene with different semantic roles for each action. Finally, we
provide a set of baseline algorithms for this task and analyze error modes
providing directions for future work. \end{abstract}

\section{Introduction}
Current state of the art on action recognition consists of classifying a video
clip containing the action, or marking a bounding box around the approximate
location of the agent doing the action. Most current action recognition
datasets classify each person into doing one of $k$ different activities and
focus on coarse activities (like `playing baseball', `cooking', `gardening').
We argue that such a coarse understanding is incomplete and a complete visual
understanding of an activity can only come when we can reason about fine
grained actions constituting each such activity (like `hitting' the ball with
a bat, `chopping' onions with a knife, `mowing' the lawn with a a lawn mower),
reason about people doing multiple such actions at the same time, and are able
to associate objects in the scene to the different semantic roles for each of these
actions. 

\begin{figure}[t] \insertA{0.48}{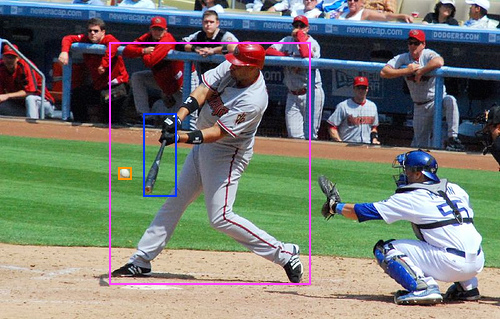} \caption{\small
\textbf{Visual Semantic Role Labeling}: We want to go beyond classifying the
action occurring in the image to being able to localize the agent, and the
objects in various semantic roles associated with the action.}
\figlabel{visual-srl} \end{figure}

\figref{visual-srl} shows our desired output. We want to go
beyond coarse activity labels such as `playing baseball', and be able to reason
about fine-grained actions such as `hitting' and detect the various
semantic roles for this action namely: the agent (pink box), the instrument
(blue box) and the object (orange box). Such an output can help us answer
various questions about the image. It tells us more about the current state of
the scene depicted in the image (association of objects in the image with each
other and with actions happening in the image), helps us better predict the
future (the ball will leave the image from the left edge of the image, the
baseball bat will swing clockwise), help us to learn commonsense about the
world (naive physics, that a bat hitting a ball impacts momentum), and in turn
help us in understanding `activities' (a baseball game is an outdoor sport
played in a field and involves hitting a round ball with a long cylindrical
bat). 

We call this problem as `Visual Semantic Role Labeling'. Semantic Role Labeling
in a Natural Language Processing context refers to labeling words in a sentence
with different semantic roles for the verb in the sentence
\cite{carreras2005introduction}. NLP research on this and related areas has
resulted in FrameNet~\cite{baker1998berkeley} and
VerbNet~\cite{KipperSchuler2006} which catalogue verbs and their semantic
roles. What is missing from such catalogues is visual grounding. Our work here
strives to achieve this grounding of verbs and their various semantic roles to
images. The set of actions we study along with the various roles are listed in
\tableref{list}.

\begin{figure*}
\centering
\renewcommand{\arraystretch}{0.5}
\setlength{\tabcolsep}{1.0pt}
\insertB{0.15}{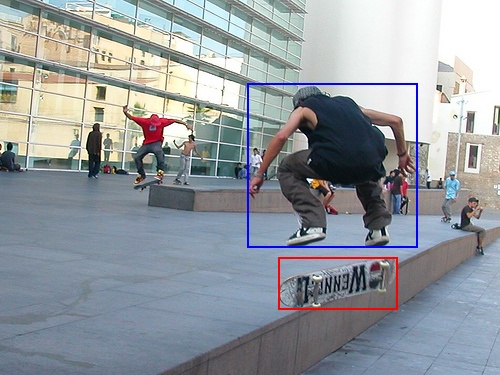}
\insertB{0.15}{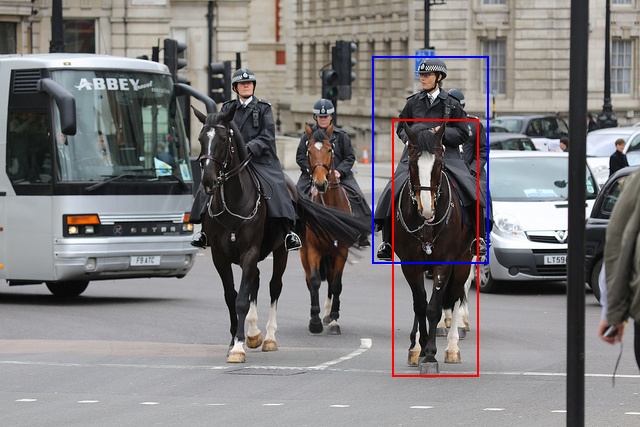}
\insertB{0.15}{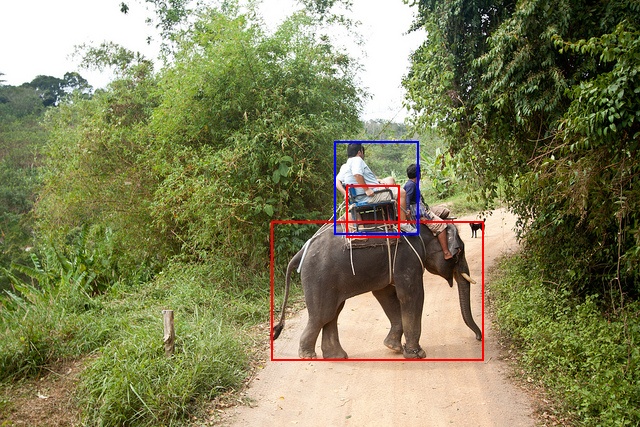}
\insertB{0.15}{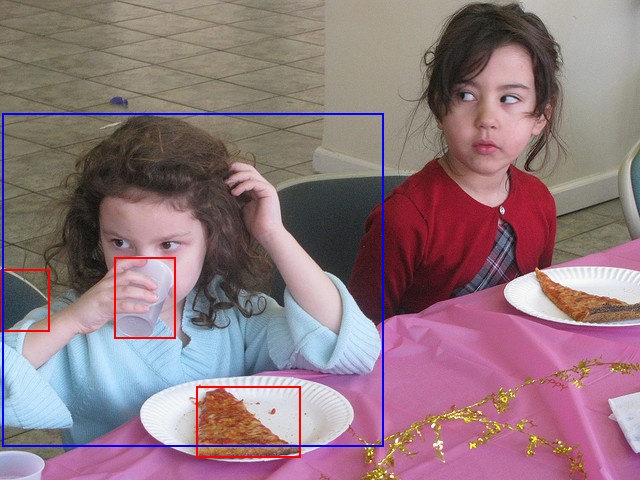}
\insertB{0.15}{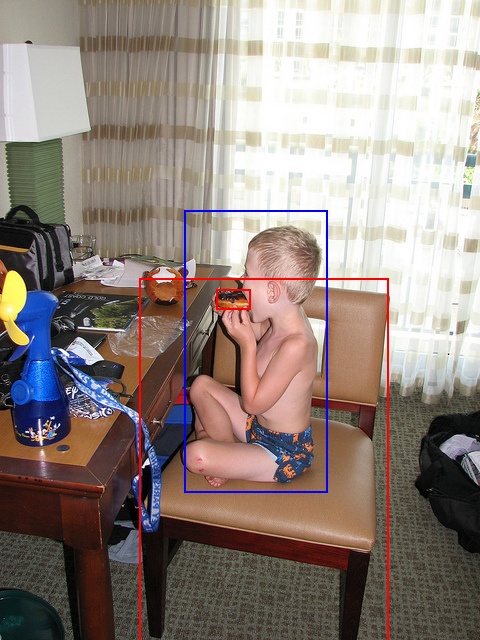}
\insertB{0.15}{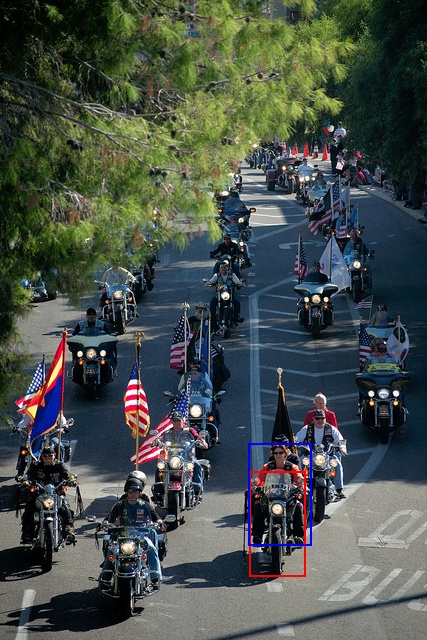} 
\insertB{0.15}{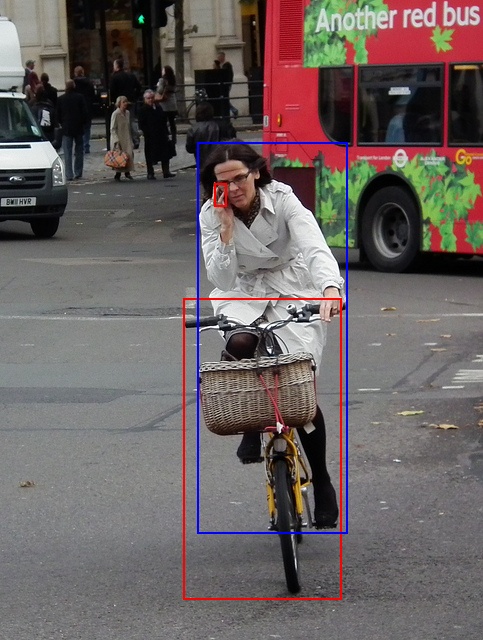} 
\insertB{0.15}{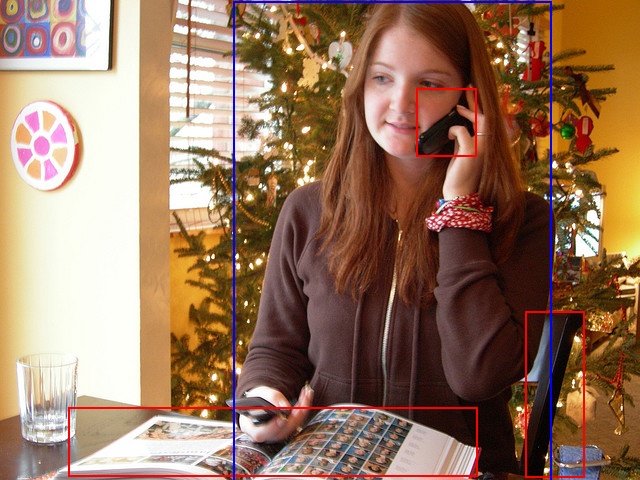}
\insertB{0.15}{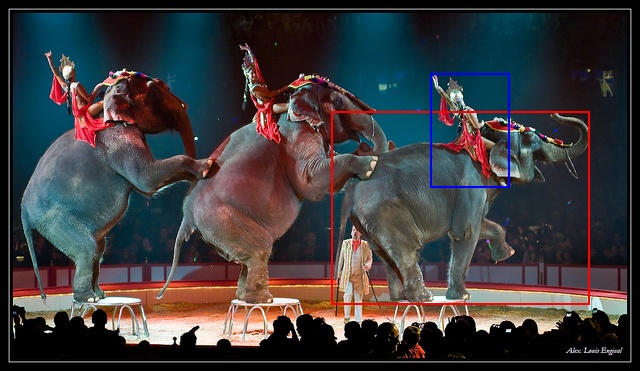}
\insertB{0.15}{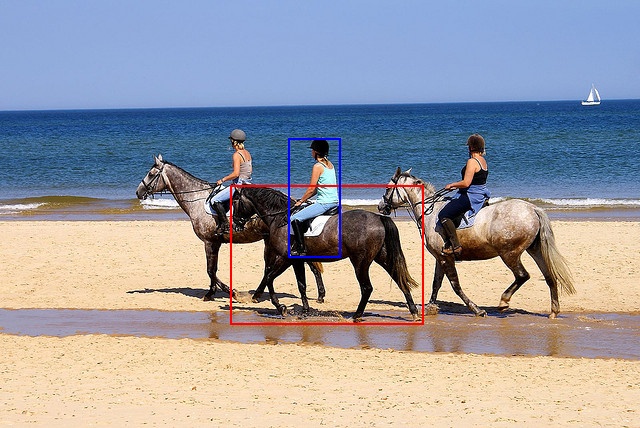}
\insertB{0.15}{}
\insertB{0.15}{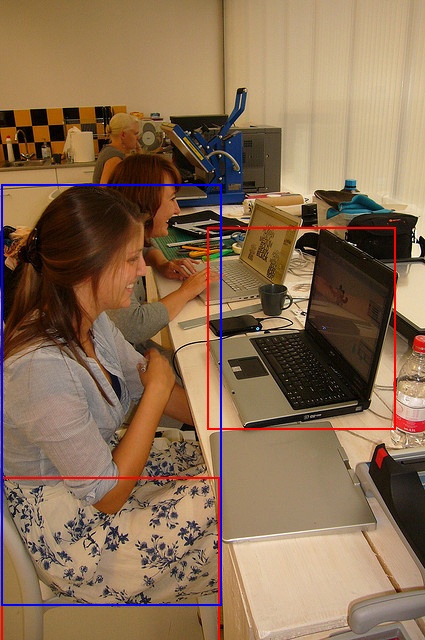}
\insertB{0.15}{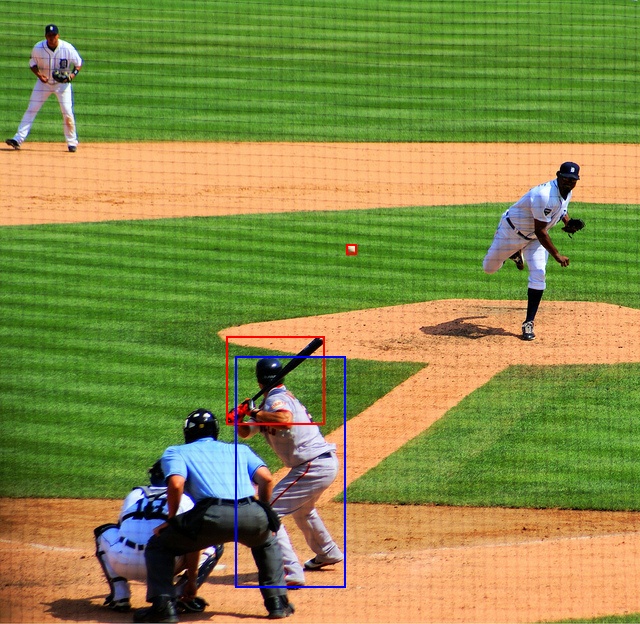}
\insertB{0.15}{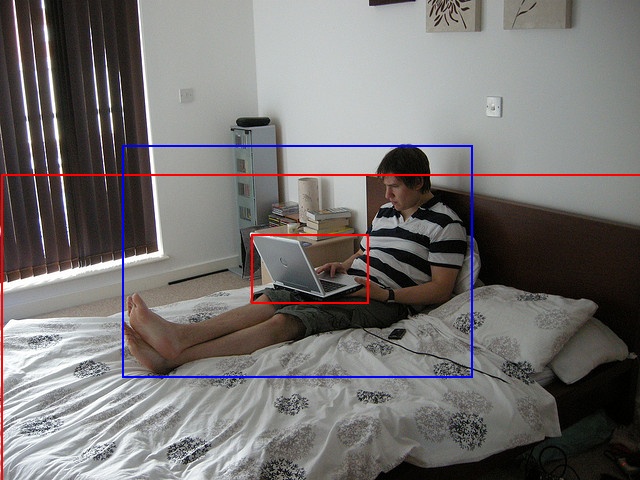}
\insertB{0.15}{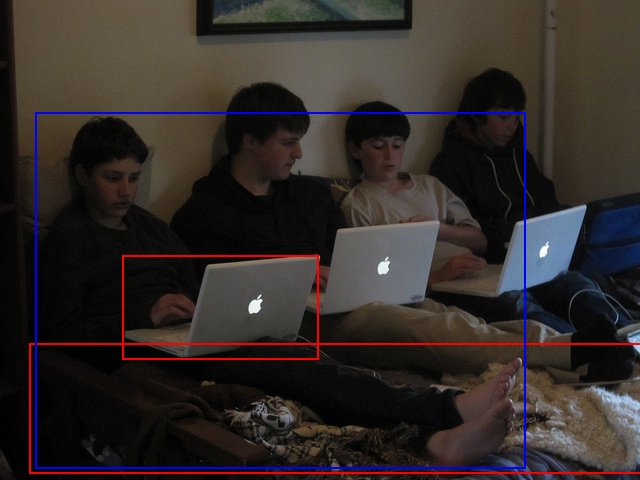}
\insertB{0.15}{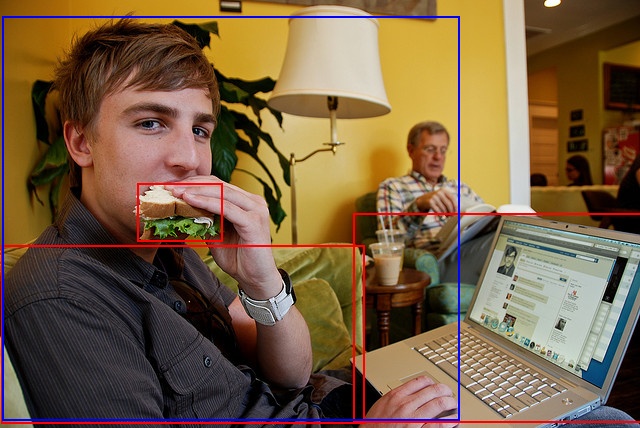} 
\insertB{0.15}{}
\insertB{0.15}{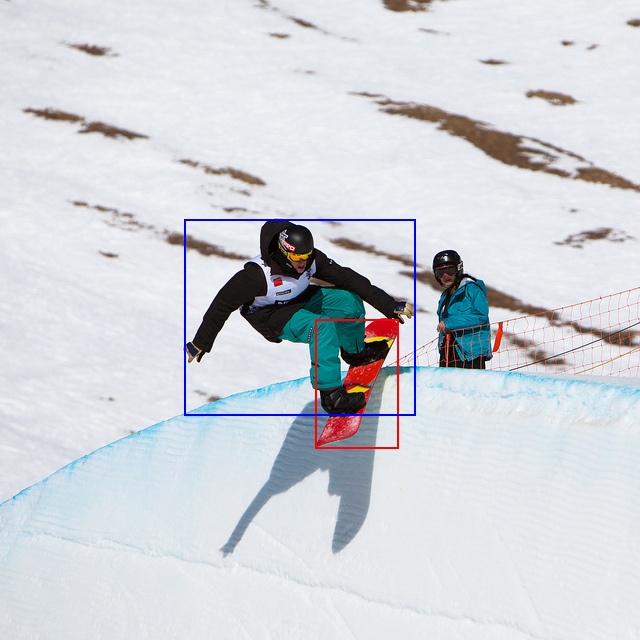}
\insertB{0.15}{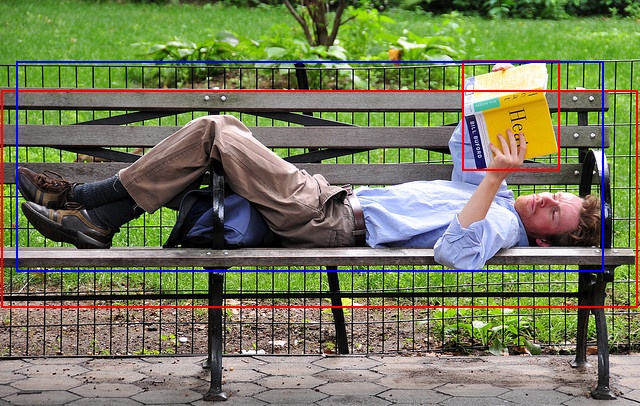}
\insertB{0.15}{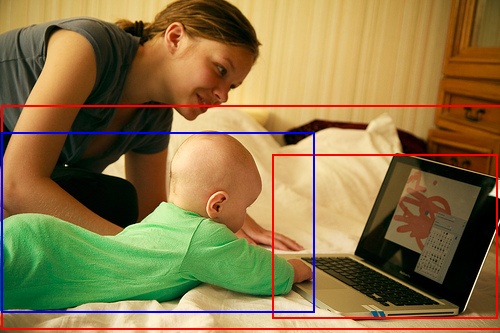}
\caption{\textbf{Visualizations of the images in the dataset}: We show examples
of annotations in the dataset. We show the agent in the blue box and objects in
various semantic roles in red. People can be doing multiple actions at the same
time. First row shows: person skateboarding, person sitting and riding on horse, person
sitting on chair and riding on elephant, person drinking from a glass and
sitting on a chair, person sitting on a chair eating a doughnut.}
\figlabel{dataset-vis}
\end{figure*}

Visual semantic role labeling is a new task which has not been studied before;
thus we start by annotating a dataset of 16K people instances with action
labels from 26 different action classes and associating objects in various
semantic roles for each person labeled with a particular action. We do this
annotation on the challenging Microsoft COCO (Common Objects in COntext)
dataset \cite{mscoco}, which contains a wide variety of objects in complex and
cluttered scenes. \figref{dataset-vis} shows some examples from our dataset.
Unlike most existing datasets which either have objects or actions labeled, as
a result of our annotation effort, \coco now has detailed action labels in
addition to the detailed object instance segmentations, and we believe will
form an interesting test bed for studying related problems. We also provide
baseline algorithms for addressing this task using CNN based object detectors,
and provide a discussion on future directions of research. 

\section{Related Work}
\seclabel{related}
There has been a lot of research in computer vision to understand activities and
actions happening in images and videos. Here we review popular action analysis
datasets, exact tasks people have studied and basic overview of techniques.

PASCAL VOC \cite{PASCAL-ijcv} is one of the popular datasets for static action
classification. The primary task here is to classify bounding box around people
instances into 9 categories. This dataset was used in the VOC challenge.
Recently, Gkioxari \etal \cite{poseactionrcnn} extended the dataset for action
detection where the task is to detect and localize people doing actions. MPII
Human Pose dataset is a more recent and challenging dataset for studying action
\cite{andriluka14cvpr}. The MPII Human Pose dataset contains 23K images
containing over 40K people with 410 different human activities. These images
come from YouTube videos and in addition to the activity label also have
extensive body part labels.  The PASCAL dataset has enabled tremendous progress
in the field of action classification, and the MPII human pose dataset has
enabled studying human pose in a very principled manner, but both these
datasets do not have annotations for the object of interaction which is the
focus of our work here. 

Gupta \etal \cite{gupta2009understanding,gupta2009observing}, Yao \etal
\cite{yao2011human,yao2012recognizing,yao2011classifying}, Prest \etal
\cite{prest2012weakly} collect and analyze the Sports, People Playing Musical
Instruments (PPMI) and the Trumpets, Bikes and Hats (TBH) datasets but the
focus of these works is on modeling human pose and object context. While Gupta
\etal and Yao \etal study the problem in supervised contexts, Prest \etal also
propose weakly supervised methods. While these methods significantly boost
performance over not using human object context and produce localization for
the object of interaction as learned by their model, they do not quantify
performance at the joint task of detecting people, classifying what they are
doing and localizing the object of interaction. Our proposed dataset will be a
natural test bed for making such quantitative measurements.

There are also a large number of video datasets for activity analysis. Some of
these study the task of full video action classification
\cite{schuldt2004recognizing,ActionsAsSpaceTimeShapes_pami07,laptev:08,marszalek09,KarpathyCVPR14,rohrbach2012database},
while some \cite{yuan2009discriminative,Jhuang:ICCV:2013,rodriguez2008action}
also study the task of detecting the agent doing the action. In particular the
J-HMDB \cite{rodriguez2008action} and UCF Sports dataset
\cite{Jhuang:ICCV:2013} are popular test beds for algorithms that study this
task \cite{actiontubes}. More recently, \cite{rohrbach15cvpr} proposed a new
video dataset where annotations come from DVS scripts. Given annotations can be
generated automatically, this will be a large dataset, but is inadequate for us
as it does not have the visual grounding which is our interest here.  

There have been a number of recent papers which generate captions for images
\cite{fangCVPR15,karpathy2014deep,kelvin2015show,ryan2014multimodal,mao2014explain,vinyals2014show,kiros2014unifying,donahue2014long,chen2014learning}.
Some of them also produce localization for various words that occur in the
sentence \cite{fangCVPR15,kelvin2015show}. 
While this maybe sufficient to generate a caption for the image, the
understanding is often limited only to the most salient action happening in the
image (based on biases of the captions that were available for training). 
A caption like `A baseball match' is completely correct for the image in
\figref{visual-srl}, but it is far from the detailed understanding we are striving
for here: an explicit action label for each person in the image along with
accurate localization for all objects in various semantic roles for the action.

\section{V-COCO Dataset} In this section, we describe the Verbs in COCO
(\vcoco) dataset. Our annotation process consisted of the following stages.
Example images from the dataset are shown in \figref{dataset-vis}.

\begin{figure}[t] 
\insertA{0.48}{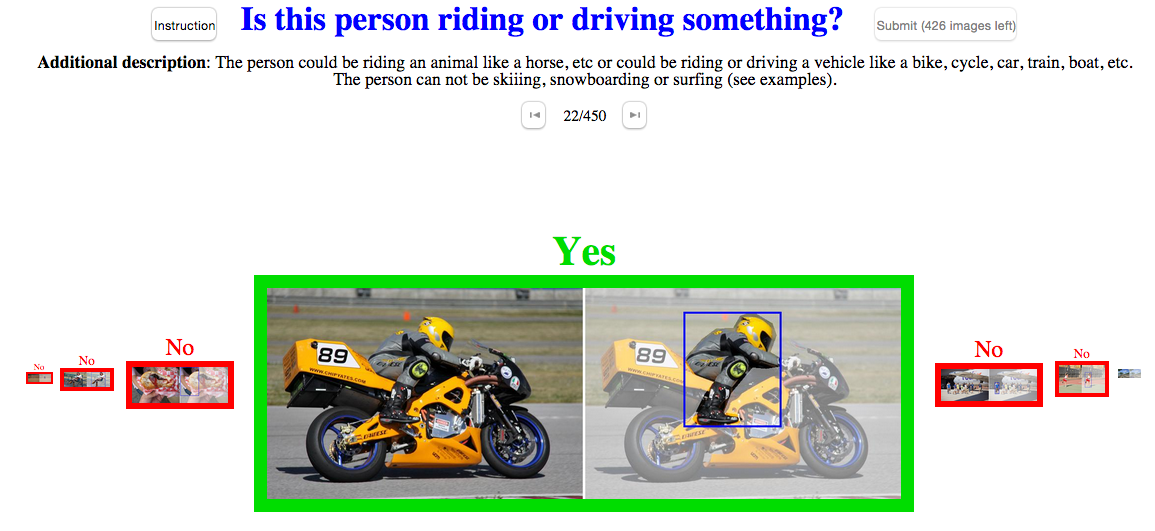} 
\insertA{0.48}{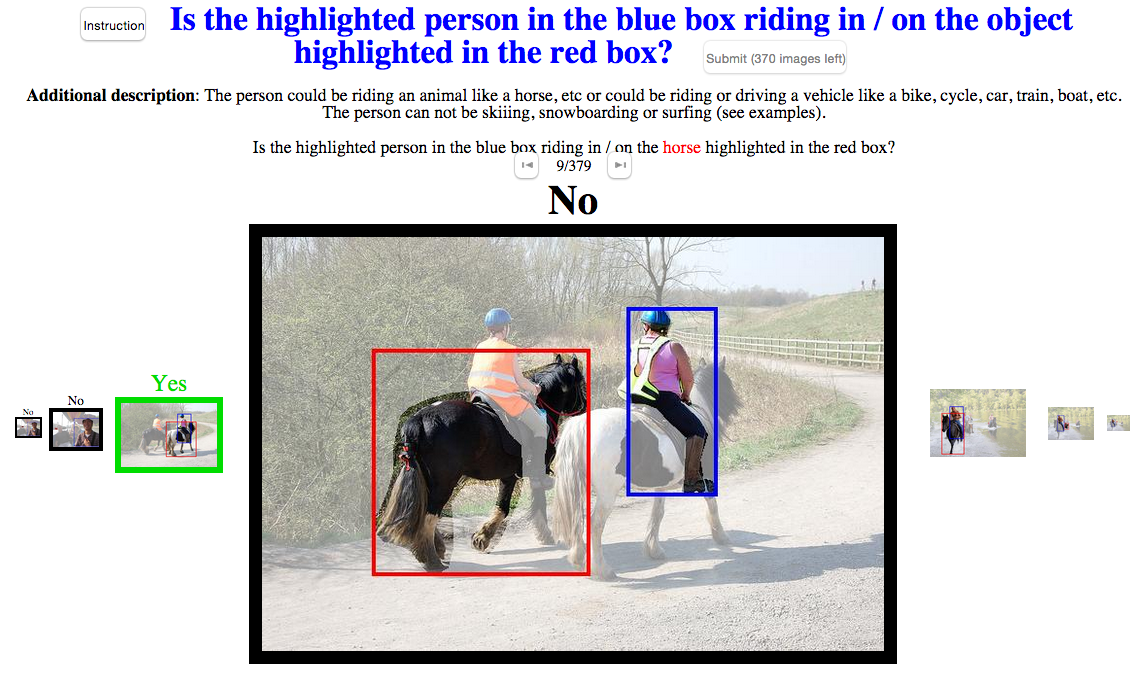} 
\caption{\small
\textbf{Interface for collecting annotations}: The top row shows the interface for
annotating the action the person is doing, and the bottom row shows the
interface for associating the roles.}
\figlabel{interface} \end{figure}

We build off the \coco dataset, for the following reasons, a) \coco is the most
challenging object detection dataset, b) it has complete annotations for 160K
images with 80 different object classes along with segmentation mask for all
objects and five human written captions for each image, c) \vcoco will get
richer if \coco gets richer \eg with additional annotations like human pose and
key points.

\textbf{Identifying verbs}
The first step is to identify a set of verbs to study. We do this in a data
driven manner. We use the captions in the \coco dataset and obtain a list of
words for which the subject is a person (we use the Stanford dependency parser
to determine the subject associated with each verb and check this subject
against a list of 62 nouns and pronouns to determine if the subject is a
person). We also obtain counts for each of these verbs (actions). Based on this
list, we manually select a set of 30 basic verbs (actions). We picked these
words with the consideration if these would be in the vocabulary of a 5-year
old child. The list of verbs is tabulated in \tableref{list}. Based on
visual inspection of images with these action words, we dropped \texttt{pick},
\texttt{place}, \texttt{give}, \texttt{take} because they were ambiguous from a
single image.

\textbf{Identifying interesting images}
With this list of verbs, the next step is to identify a set of images
containing people doing these actions. We do this independently for each verb.
We compute two scores for each image: a) does this image have a person
associated with the target verb (or its synonyms) (based on the captions for
the image), b) does this image contain objects associated with the target verb
(using the \coco object instance annotations, the list of associated objects
was picked manually). Query expansion using the set of objects associated with
the target verb was necessary to obtain enough examples.

We sum these scores to obtain a ranked list of images for each verb, and
consider the top 8000 images independently for each verb (in case the above
two scores do not yield enough images we take additional images that contain
people).  We then use \amt to obtain annotations for people in these 8000
images (details on the mechanical turk annotation procedure are provided in
\secref{amt}). We thus obtain a set of positive instances for each verb. The
next step is to come up with a common set of images across all action
categories. We do this by solving an integer program. This step of obtaining
annotations separately for each action and then merging positive instances into
a common pool of images was important to get enough examples for each action
class. 

\textbf{Salient People} Given this task requires detailed reasoning (consider
localizing the spoon and fork being used to \vb{eat}), instead of working with
all people in the image, we work with people instances which have sufficient
pixel area in the image. In addition, we also discard all people with pixel
area less than half the pixel area of the largest person in the image. This
helps with images which have a lot of by-standers (who may not be doing
anything interesting). Doing this speeds up the annotation process
significantly, and allows us to use the annotation effort more effectively.
Note that we can still study the \vsrl problem in a detection setting. Given
the complete annotations in \coco, even if we don't know the action that a non
salient person is doing, we still know its location and appropriately adjust
the training and evaluation procedures to take this into account.

\textbf{Annotating salient people with all action labels} 
Given this set of images we annotate all `salient people' in these images with
binary label for each action category. We again use \amt for obtaining these
annotations but obtain annotations from 5 different workers for each person for
each action.

\textbf{Annotating object in various roles}
Finally, we obtain annotations for objects in various roles for each action. We
first enumerate the various roles for each verb, and identify object
categories that are appropriate for these roles (see \tableref{list}). For
each positively annotated person (with 3 or more positive votes from the
previous stage) we obtain YES/NO annotation for questions of the form:
`{\small Is the \textit{person} in the blue box \textit{holding} the
\textit{banana} in the red box?}'

\textbf{Splits}
To minimize any differences in statistics between the different splits of the
data, we combined 40K images from the \coco training set with the \coco
validation set and obtained annotations on this joint set. After the
annotation, we construct 3 splits for the \vcoco dataset: the images coming
from the validation set in \coco were put into the \vcoco \textit{test} set,
the rest of the images were split into \vcoco \textit{train} and \textit{val}
sets. 

\subsection{Annotation Procedure} \seclabel{amt} In this section, we describe
the Amazon Mechanical Turk (\amt) \cite{AMT} annotation procedure that we use
during the various stages of dataset annotation.

We follow insights from Zhou \etal \cite{zhou2014places} and use their
annotation interface. We frame each annotation task as a binary {\small
{YES}/{NO}} task. This has the following advantages: the user interface for
such a task is simple, it is easy to insert test questions, it is easy to
asses consensus, and such a task ends up getting done faster on \amt.

\textbf{User Interface} We use the interface from Zhou \etal
\cite{zhou2014places}. The interface is shown in \figref{interface}. We show
two images: the original image on the left, and the image highlighting the
person being annotated on the right. All images were marked with a default NO
answer and the turker flips the answer using a key press. We inserted test
questions to prevent spamming (see below) and filtered out inaccurate turkers.
We composed HITs (Human Intelligence Tasks) with 450 questions (including 50
test questions) of the form: `{\small Is the person highlighted in the blue box
\textit{holding} something?}' In a given HIT, the action was kept fixed. On
average turkers spent 15 minutes per HIT, although this varied from action to
action. For annotating the roles, an additional box was highlighted
corresponding to the object, and the question was changed appropriately. 

\textbf{Test Questions} We had to insert test questions to ensure
reliability of turker annotations. We inserted two sets of test questions. The
first set was used to determine accuracy at the time of submission and this
prevented turkers from submitting answers if their accuracy was too low (below
90\%). The second set of questions were used to guard from turkers who hacked
the client side testing to submit incorrect results (surprisingly, we did find
turkers who did this). HITs for which the average accuracy on the test set was
lower than 95\% were relaunched. The set of test questions were bootstrapped
from the annotation process, we started with a small set of hand labeled test
questions, and enriched that set based on annotations obtained on a small set
of images. We manually inspected the annotations and augmented the test set to
penalize common mistakes (skiing \vs snowboarding) and excluded ambiguous
examples.

\subsection{Dataset Statistics}

\renewcommand{\arraystretch}{1.3}
\begin{table}
\caption{\textbf{List of actions in \vcoco.} We list the different actions, number of
semantic roles associated with the action, number of examples for each action,
the different roles associated with each action along with their counts and the
different objects that can be take each role. Annotations for cells marked with * are
currently underway.} \vspace{2mm}
\tablelabel{list}
\scalebox{0.6}{
\begin{tabular}{>{\raggedright}p{0.08\textwidth}crrrp{0.35\textwidth}r} \toprule
           Action        & Roles & \#    & Role    & \#    & Objects in role                                                                                                    & \\ \midrule
            carry        & 1               & 970  & \object & *    &                                                                                                                    & \\
            catch        & 1               & 559  & \object & 457  & sports ball, frisbee,                                                                                              & \\
              cut        & 2               & 569  & \instr  & 477  & scissors, fork, knife,                                                                                             & \\
                         &                 &      & \object & *    &                                                                                                                    & \\
            drink        & 1               & 215  & \instr  & 203  & wine glass, bottle, cup, bowl,                                                                                     & \\
              eat        & 2               & 1198 & \object & 737  & banana, apple, sandwich, orange, carrot, broccoli, hot dog, pizza, cake, donut,                                    & \\
                         &                 &      & \instr  & *    &                                                                                                                    & \\
              hit        & 2               & 716  & \instr  & 657  & tennis racket, baseball bat,                                                                                       & \\
                         &                 &      & \object & 454  & sports ball                                                                                                        & \\
             hold        & 1               & 7609 & \object & *    &                                                                                                                    & \\
             jump        & 1               & 1335 & \instr  & 891  & snowboard, skis, skateboard, surfboard,                                                                            & \\
             kick        & 1               & 322  & \object & 297  & sports ball,                                                                                                       & \\
              lay        & 1               & 858  & \instr  & 513  & bench, dining table, toilet, bed, couch, chair,                                                                    & \\
             look        & 1               & 7172 & \object & *    &                                                                                                                    & \\
            point        & 1               & 69   & \object & *    &                                                                                                                    & \\
             read        & 1               & 227  & \object & 172  & book,                                                                                                              & \\
             ride        & 1               & 1044 & \instr  & 950  & bicycle, motorcycle, bus, truck, boat, train, airplane, car, horse, elephant,                                      & \\
              run        & 0               & 1309 & -       &   -  &                                                                                                                    & \\
              sit        & 1               & 3905 & \instr  & 2161 & bicycle, motorcycle, horse, elephant, bench, chair, couch, bed, toilet, dining table, suitcase, handbag, backpack, & \\
       skateboard        & 1               & 906  & \instr  & 869  & skateboard,                                                                                                        & \\
              ski        & 1               & 924  & \instr  & 797  & skis,                                                                                                              & \\
            smile        & 0               & 2960 &    -    &  -   &                                                                                                                    & \\
        snowboard        & 1               & 665  & \instr  & 628  & snowboard,                                                                                                         & \\
            stand        & 0               & 8716 &   -     &  -   &                                                                                                                    & \\
             surf        & 1               & 984  & \instr  & 949  & surfboard,                                                                                                         & \\
    talk on phone        & 1               & 639  & \instr  & 538  & cell phone,                                                                                                        & \\
            throw        & 1               & 544  & \object & 475  & sports ball, frisbee,                                                                                              & \\
             walk        & 0               & 1253 &     -   &  -   &                                                                                                                    & \\
 work on computer        & 1               & 868  & \instr  & 773  & laptop,                                                                                                            & \\
\bottomrule
\end{tabular}}
\end{table}

\begin{table}
\setlength{\tabcolsep}{1.2pt}
\begin{center}
\caption{List and counts of actions that co-occur in \vcoco.}
\tablelabel{co-occur-stats}
\scriptsize{
\begin{tabular}{>{\textbf}lp{0.13\textwidth}lp{0.13\textwidth}lp{0.13\textwidth}} \toprule
                 597 & look, stand                 &     &                                   & 411 & carry, hold, stand, walk   \\
                 340 & hold, stand, ski            &     &                                   & 329 & ride, sit   \\
                 324 & look, sit, work on computer & 302 & look, jump, skateboard            & 296 & hold, ride, sit   \\
                 280 & look, surf                  & 269 & hold, stand                       & 269 & hold, look, stand   \\
                 259 & hold, smile, stand          & 253 & stand, walk                       & 253 & hold, sit, eat   \\
                 238 & smile, stand                & 230 & look, run, stand                  & 209 & hold, look, stand, hit   \\
                     &                             & 193 & hold, smile, stand, ski           & 189 & look, sit   \\
                 189 & look, run, stand, kick      & 183 & smile, sit                        &     &        \\
                 160 & look, stand, surf           & 159 & hold, look, sit, eat              & 152 & hold, stand, talk on phone   \\
                 150 & stand, snowboard            & 140 & hold, look, smile, stand, cut     &     &        \\
                 129 & hold, stand, throw          & 128 & look, stand, jump, skateboard     & 127 & hold, smile, sit, eat   \\
                 124 & hold, look, stand, cut      & 121 & hold, look, sit, work on computer & 117 & hold, stand, hit   \\
                 115 & look, jump, snowboard       & 115 & look, stand, skateboard           & 113 & stand, surf   \\
                 107 & hold, look, run, stand, hit & 105 & look, stand, throw                & 104 & hold, look, stand, ski   \\
\bottomrule
\end{tabular}}
\end{center}
\end{table}

In this section we list statistics on the dataset. The \vcoco
dataset contains a total of 10346 images containg 16199 people
instances. Each annotated person has binary labels for 26 different actions.
The set of actions and the semantic roles associated with each action are
listed in \tableref{list}. \tableref{list} also lists the number of positive
examples for each action, the set of object categories for various roles for each
ation, the number of instances with annotations for the object of interaction.

We split the \vcoco dataset into a \train, \val and \test split. The \train and
\val splits come from the \coco \train set while the \test set comes from the
\val set. Number of images and annotated people instances in each of these
splits are tabulated in \tableref{stat-splits}.

Note that all images in \vcoco inherits all the annotations from the \coco
dataset \cite{mscoco}, including bounding boxes for non-salient people, crowd regions,
allowing us to study all tasks in a detection setting. Moreover, each image
also has annotations for 80 object categories which can be used to study the
role of context in such tasks.

\figref{hists} (left) shows the distribution of the number of people instances
in each image. Unlike past datasets which mostly have only one annotated person
per image, the \vcoco dataset has a large number of images with more than one
person. On average these have 1.57 people annotated with action labels per
image. There are about 2000 images with two, and 800 images with three people
annotated. 

\figref{hists} (right) shows a distribution of the number of different actions
a person is doing in \vcoco. Unlike past datasets where each person can only be
doing one action, people in \vcoco do on average 2.87 actions at the same time.
\tableref{co-occur-stats} lists the set of actions which co-occur more than 100
times along with their counts. We also analyse human agreement for different
actions to quantify the ambiguity in labeling actions from a single image by
benchmarking annotations from one turker with annotations from the other
turkers for each HIT for each action and produce points on precision and recall
plot. \figref{human-agreement} presents these plots for the \vb{walk}, \vb{run}
and \vb{surf} actions. We can see that there is high human agreement for
actions like \vb{surf}, where as there is lower human agreement for verbs like
\vb{walk} and \vb{run}, as expected.

\begin{figure}
\centering
\insertA{0.23}{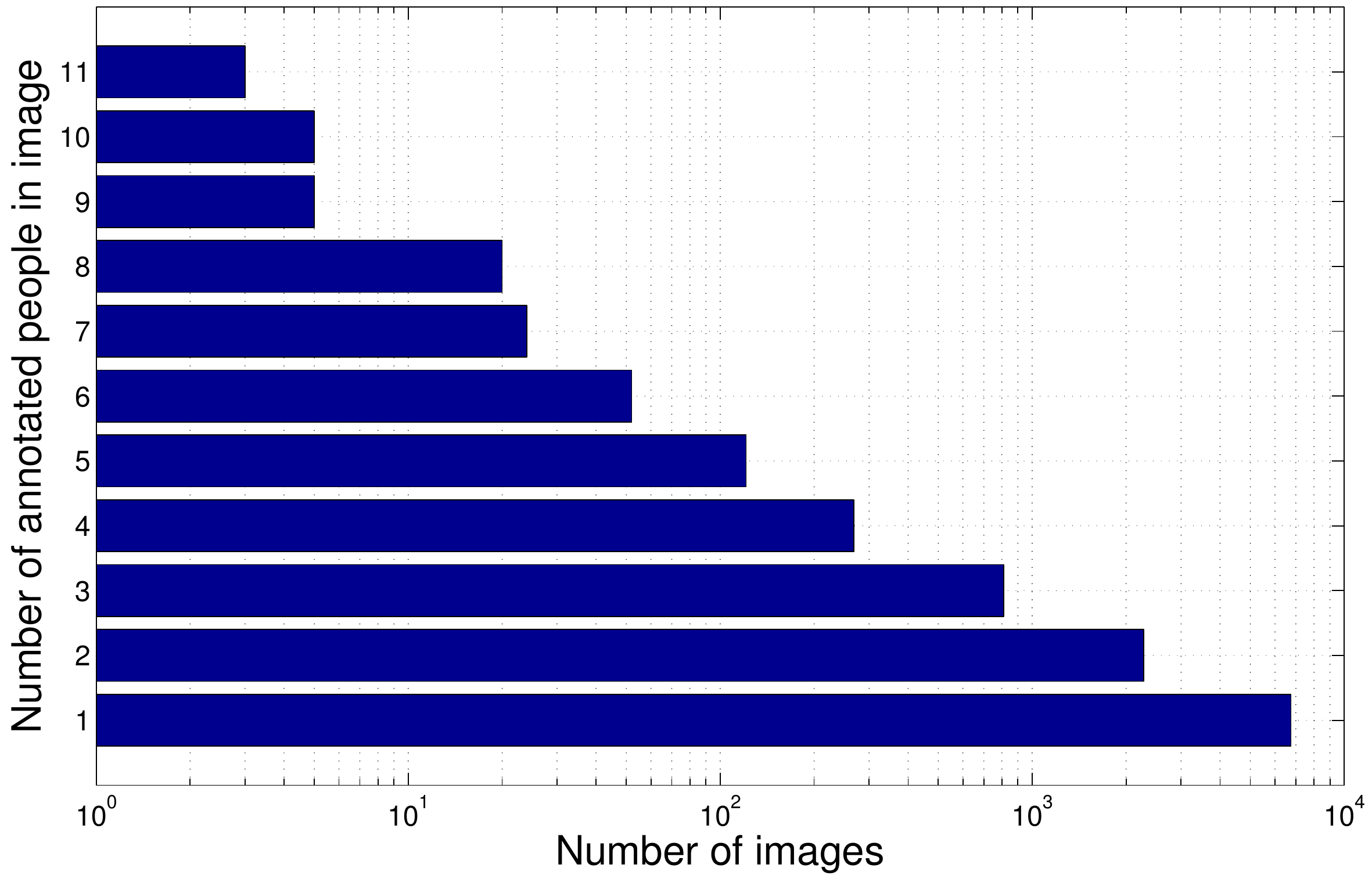}
\insertA{0.23}{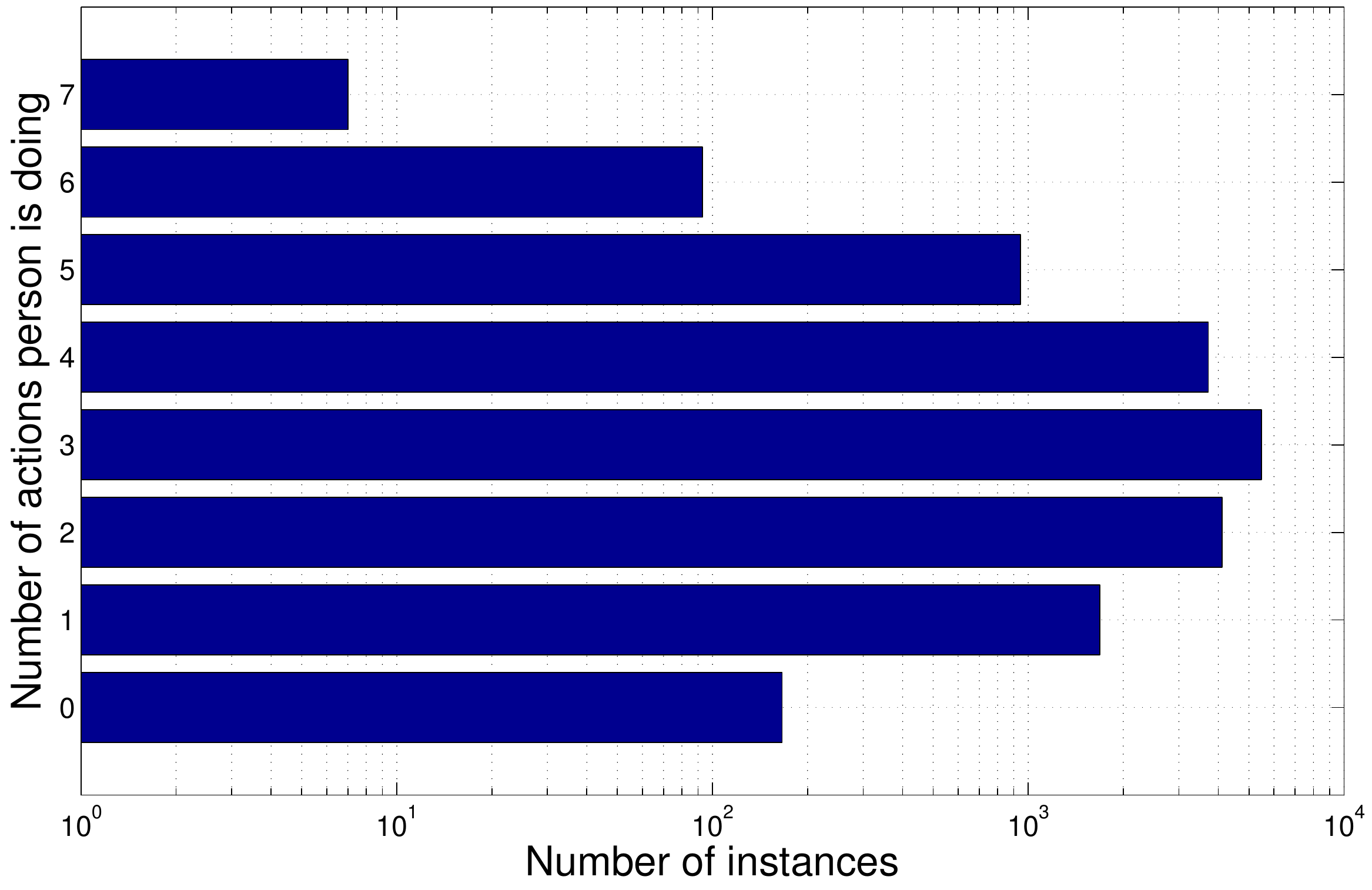} 
\caption{\textbf{Statistics on \vcoco}: The bar plot on left shows the
distribution of the number of annotated people per image. The bar plot on right
shows the distribution of the number of actions a person is doing. Note that
X-axis is on $\log$ scale.}
\figlabel{hists}
\end{figure}

\begin{figure}
  \insertA{0.1492}{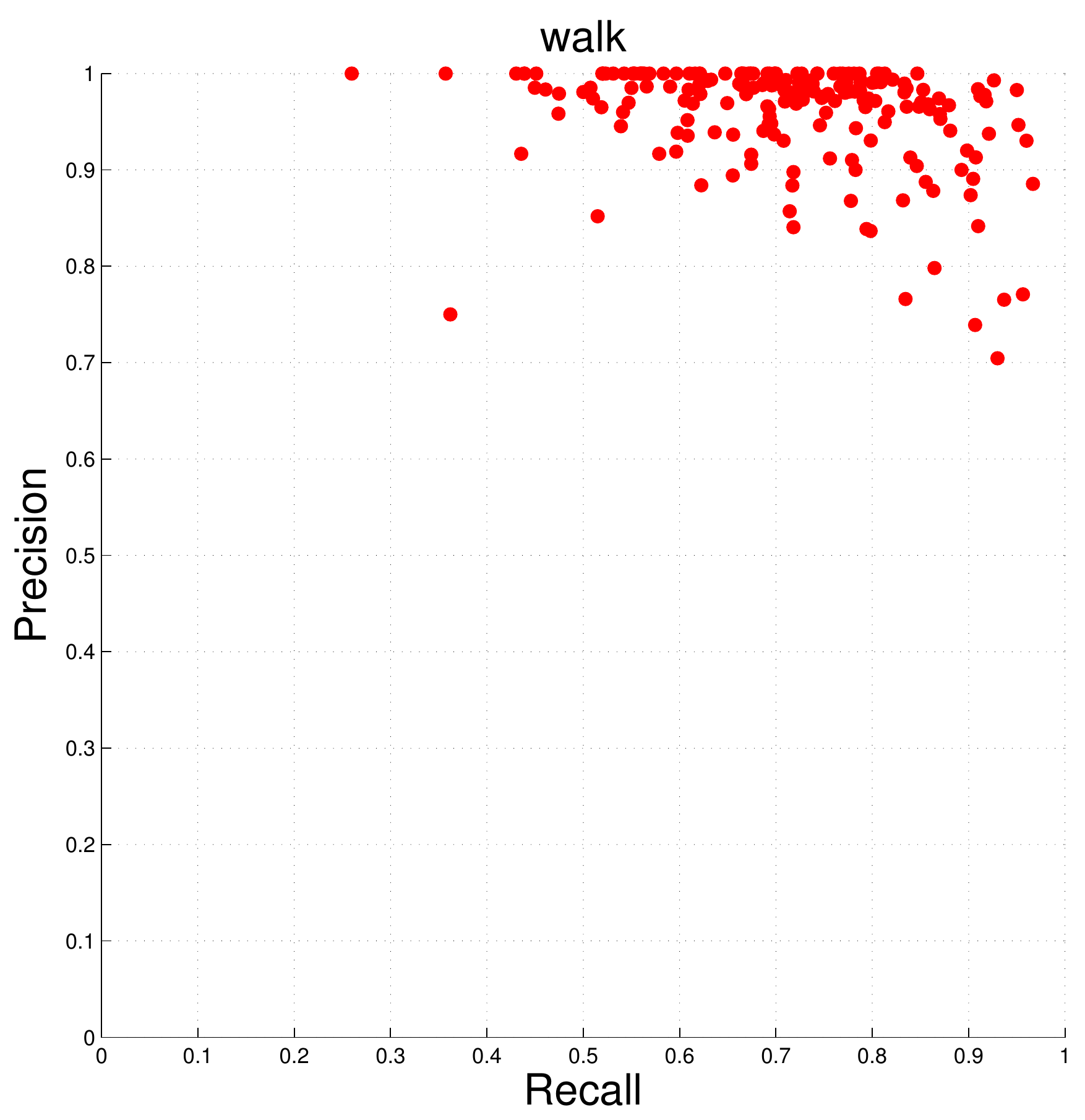}
  \insertA{0.1492}{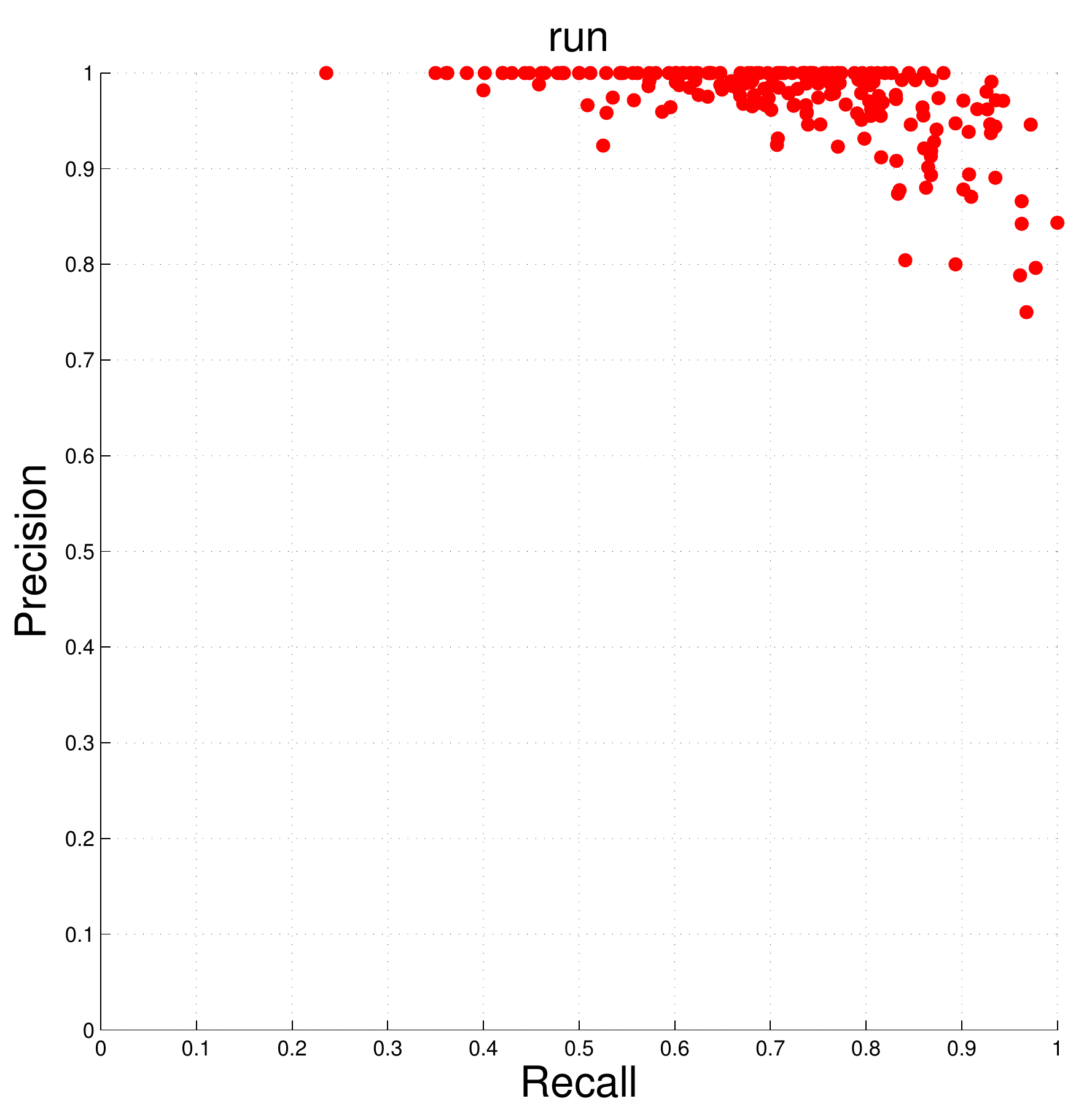}
  \insertA{0.1492}{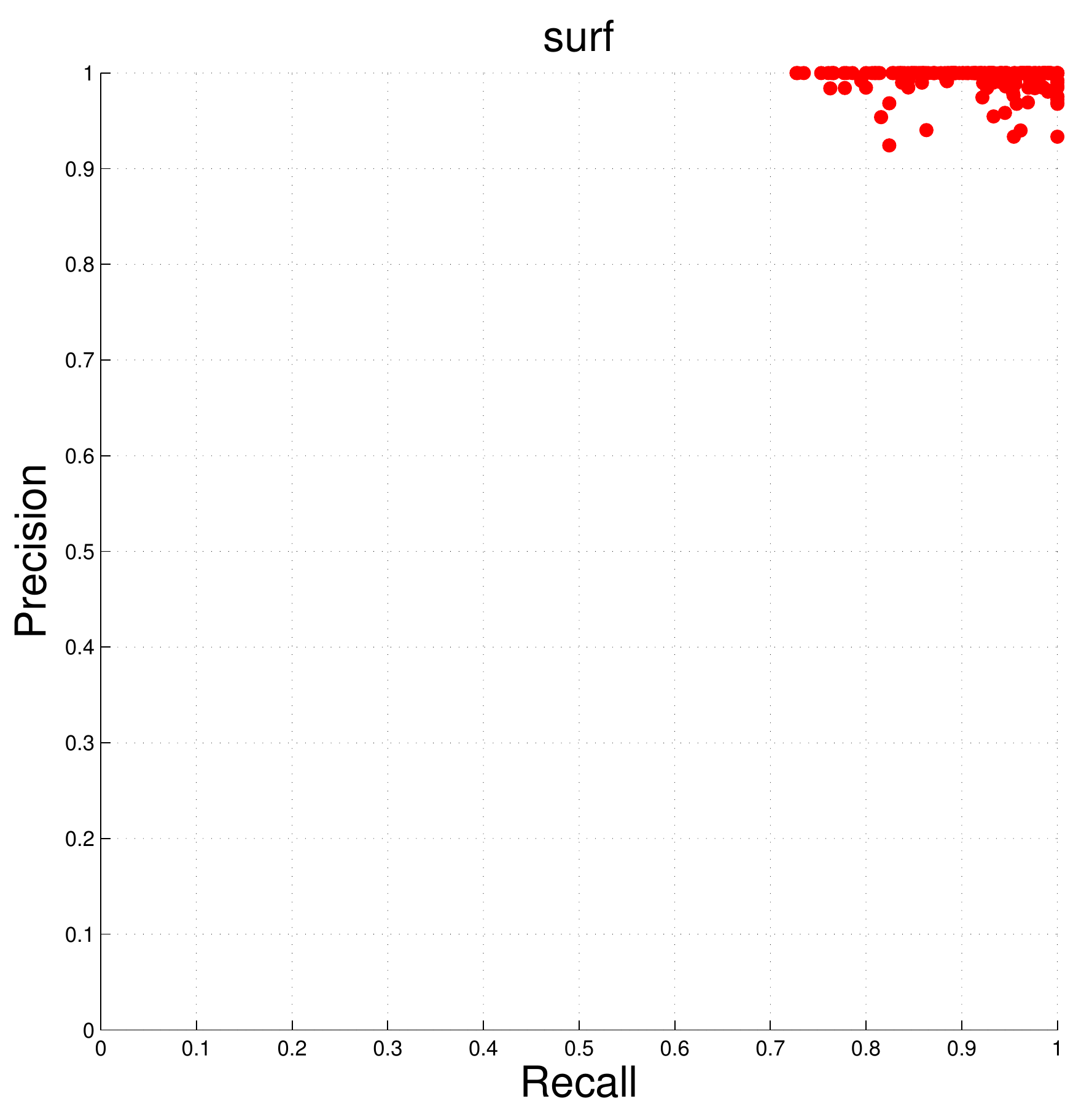}
\caption{Human Agreement}
\figlabel{human-agreement}
\end{figure}

\renewcommand{\arraystretch}{1.2}
\begin{table}
\caption{Statistics of various splits of \vcoco.}
\tablelabel{stat-splits}
\begin{center}
\scalebox{0.8}{
\begin{tabular}{lcccc}
\toprule
  & \train & \val & \test & \textit{all} \\ \midrule
Number of Image & 2533 & 2867 & 4946 & 10346 \\ 
Number of People Instance & 3932 & 4499 & 7768 & 16199 \\  \bottomrule
\end{tabular}}
\end{center}
\end{table}

\subsection{Tasks and Metrics}
These annotations enable us to study a variety of new fine grained tasks about
action understanding which have not been studied before. We describe these
tasks below.

\textbf{Agent Detection} The agent detection task is to detect instances of
people engaging in a particular action. We use the standard average precision
metric as used for PASCAL VOC object detection \cite{PASCAL-ijcv} to measure
performance at this task - people labeled positively with the action category
are treated as positive, un-annotated non-salient people are marked as difficult. 

\textbf{Role Detection} The role detection task is to detect the agent and
the objects in the various roles for the action. An algorithm produces as
output bounding boxes for the locations of the agent and each semantic role. A
detection is correct if the location of the agent and each role is correct
(correctness is measured using bounding box overlap as is standard). As an
example, consider the role detection task for the action class `hold'. An
algorithm will have to produce as output a bounding box for the person
`holding', and the object being `held', and both these boxes must be correct
for this detection to be correct. We follow the same precision recall 
philosophy and use average precision as the metric.

\section{Methods}
\seclabel{method}
In this section, we describe the baseline approaches we investigated for
studying this task. As a first step, we train object detectors for the 80
different classes in the \coco dataset. We use \rcnn \cite{girshickCVPR14} to
train these detectors and use the 16-layer CNN from Simonyan and Zisserman
\cite{simonyan2014very} (we denote this as \vgg). This \cnn has been shown to be
very effective at a variety of tasks like object detection
\cite{girshickCVPR14}, image captioning \cite{fangCVPR15}, action
classification \cite{pascal_leaderboard}. We finetune this detector using the
fast version of R-CNN \cite{fastrcnn} and train on 77K images from the \coco
train split (we hold out the 5K \vcoco \train and \val images). We use the
precomputed MCG bounding boxes from \cite{pont2015multiscale}. 

\paragraph{Agent detection model} Our model for agent detection starts by
detecting people, and then classifies the detected people into different action
categories. We train this classification model using MCG bounding boxes which
have an intersection over union of more than 0.5 with the ground truth bounding
box for the person. Since each person can be doing multiple actions at the same
time, we frame this as a multi-label classification problem, and finetune the
\vgg representation for this task. We denote this model as $A$. 

At test time, each person detection (after non-maximum suppression), is scored
with classifiers for different actions, to obtain a probability for each
action. These action probabilities are multiplied with the probability from the
person detector to obtain the final score for each action class. 

\paragraph{Regression to bounding box for the role} Our first attempt to
localize the object in semantic roles associated with an action involves
training a regression model to regress to the location of the semantic role.
This regression is done in the coordinate frame of the detected agent (detected
using model $A$ as described above). We use the following 4 regression targets
\cite{girshickCVPR14}. $(\bar{x}_{t}, \bar{y}_{t})$ denotes the center of the
target box $t$, $(\bar{x}_o, \bar{y}_{o})$ denotes the center of the detected
person box $o$, and $(w_{t}, h_{t})$, $(w_o, h_o)$ are the width and height of
the target and person box.
\begin{eqnarray}
\delta(t,o) = \left(\frac{\bar{x}_{t} - \bar{x}_{o}}{w_{o}}, \frac{\bar{y}_{t} -
\bar{y}_{o}}{h_{o}},
\log\left(\frac{{w}_{t}}{w_{o}}\right), \log\left(\frac{{h}_{t}}{h_{o}}\right) \right)
\eqlabel{delta}
\end{eqnarray}
We denote this model as $B$.

\paragraph{Using Object Detectors} Our second method for localizing these
objects uses object detectors for the categories that can be a part of the
semantic role as described in \tableref{list}. We start with the detected agent
(using model $A$ above) and for each detected agent attach the highest scoring
box according to the following score function: 
\begin{eqnarray}
P_D\left(\delta(t_{c},o)\right) \times sc_{c}(t_{c})
\end{eqnarray}
where $o$ refers to the box of the detected agent, box $t_c$ comes from all
detection boxes for the relevant object categories $c \in \mathcal{C}$ for that
action class, and $sc_c(t_c)$ refers to the detection probability for object
category $c$ for box $t_{c}$. $P_D$ is the probability distribution of
deformations $\delta$ computed from the training set, using the annotated agent
and role boxes. We model this probability distribution using a Gaussian. The
detection probabilities for different object categories $c \in \mathcal{C}$ are
already calibrated using the softmax in the \fastrcnn training \cite{fastrcnn}.
We denote this model as $C$. 

\section{Experiments}
\seclabel{exp}
\renewcommand{\arraystretch}{1.3}
\begin{table}
\centering
\caption{Performance on actions in \vcoco. We report the recall for MCG
candidates for objects that are part of different semantic roles for each
action, AP for agent detection and role detection for 4 baselines using VGG CNN
(with and without finetuning for this task). See \secref{exp} for more
details.}
\vspace{2mm}
\tablelabel{ap}
\scalebox{0.6}{
\begin{tabular}{>{\raggedleft}p{0.08\textwidth}crrrrrrrrrr} \toprule
           Action & Role & \multicolumn{3}{c}{MCG Recall} & & \multicolumn{5}{c}{Average Precision}\\ 
           \cmidrule(r){3-5} \cmidrule(r){7-11} 
                   &          &      &        &        &  & $A$   & $B_0$ & $B$  & $C_0$ & $C$  & \\
                   &          & mean & R[0.5] & R[0.7] &  & agent & role  & role & role  & role & \\ \midrule
            carry  & \object* &      &        &        &  & 54.2  &       &      &       &      & \\
            catch  & \object  & 73.7 & 91.6   & 67.2   &  & 41.4  & 1.1   & 1.2  & 24.1  & 22.5 & \\
              cut  & \instr   & 58.6 & 61.4   & 30.7   &  & 44.5  & 1.6   & 2.3  & 4.6   & 3.9  & \\
                   & \object* &      &        &        &  &       &       &      &       &      & \\
            drink  & \instr   & 69.5 & 82.1   & 58.2   &  & 25.1  & 0.3   & 0.7  & 3.1   & 6.4  & \\
              eat  & \object  & 84.4 & 97.8   & 89.7   &  & 70.2  & 8.0   & 11.0 & 37.0  & 46.2 & \\
                   & \instr*  &      &        &        &  &       &       &      &       &      & \\
              hit  & \instr   & 72.0 & 88.1   & 60.5   &  & 82.6  & 0.2   & 0.7  & 31.0  & 31.0 & \\
                   & \object  & 62.0 & 73.8   & 53.3   &  &       & 11.3  & 11.8 & 41.3  & 44.6 & \\
             hold  & \object* &      &        &        &  & 73.4  &       &      &       &      & \\
             jump  & \instr   & 76.0 & 88.7   & 68.7   &  & 69.2  & 4.0   & 17.0 & 33.9  & 35.3 & \\
             kick  & \object  & 82.7 & 100.0  & 94.4   &  & 61.6  & 0.3   & 0.8  & 48.8  & 48.3 & \\
              lay  & \instr   & 94.6 & 100.0  & 97.7   &  & 39.3  & 19.9  & 28.0 & 32.8  & 34.3 & \\
             look  & \object* &      &        &        &  & 65.0  &       &      &       &      & \\
            point  & \object* &      &        &        &  & 1.4   &       &      &       &      & \\
             read  & \object  & 83.5 & 96.2   & 82.7   &  & 10.6  & 0.9   & 2.1  & 2.2   & 4.7  & \\
             ride  & \instr   & 84.7 & 99.1   & 87.9   &  & 45.4  & 1.2   & 9.7  & 12.5  & 27.6 & \\
              run  & -        &      &        &        &  & 59.7  &       &      &       &      & \\
              sit  & \instr   & 82.2 & 94.7   & 82.6   &  & 64.1  & 20.0  & 22.3 & 24.3  & 29.2 & \\
       skateboard  & \instr   & 73.2 & 87.3   & 63.7   &  & 83.7  & 3.0   & 12.2 & 32.7  & 40.2 & \\
              ski  & \instr   & 49.1 & 46.5   & 20.9   &  & 81.9  & 4.9   & 5.5  & 5.9   & 8.2  & \\
            smile  & -        &      &        &        &  & 61.9  &       &      &       &      & \\
        snowboard  & \instr   & 67.8 & 75.1   & 51.7   &  & 75.8  & 4.3   & 13.6 & 20.2  & 28.1 & \\
            stand  & -        &      &        &        &  & 81.0  &       &      &       &      & \\
             surf  & \instr   & 66.7 & 76.0   & 53.2   &  & 94.0  & 1.5   & 4.8  & 28.1  & 27.3 & \\
    talk on phone  & \instr   & 59.9 & 69.3   & 37.3   &  & 46.6  & 1.1   & 0.6  & 5.8   & 5.8  & \\
            throw  & \object  & 72.5 & 88.0   & 73.6   &  & 50.1  & 0.4   & 0.5  & 25.7  & 25.4 & \\
             walk  & -        &      &        &        &  & 56.3  &       &      &       &      & \\
 work on computer  & \object  & 85.6 & 98.6   & 88.5   &  & 56.9  & 1.4   & 4.9  & 29.8  & 32.3 & \\ \midrule
             mean  &          & 73.6 & 85.0   & 66.4   &  & 57.5  & 4.5   & 7.9  & 23.4  & 26.4 & \\ \bottomrule
\end{tabular}}
\end{table}

\begin{figure*}
\centering
\renewcommand{\arraystretch}{0.3}
\setlength{\tabcolsep}{1.0pt}
\begin{tabular}{p{0.02\textwidth}>{\raggedright\arraybackslash}p{0.98\textwidth}}
\vertical{Correct} &
\insertB{0.13}{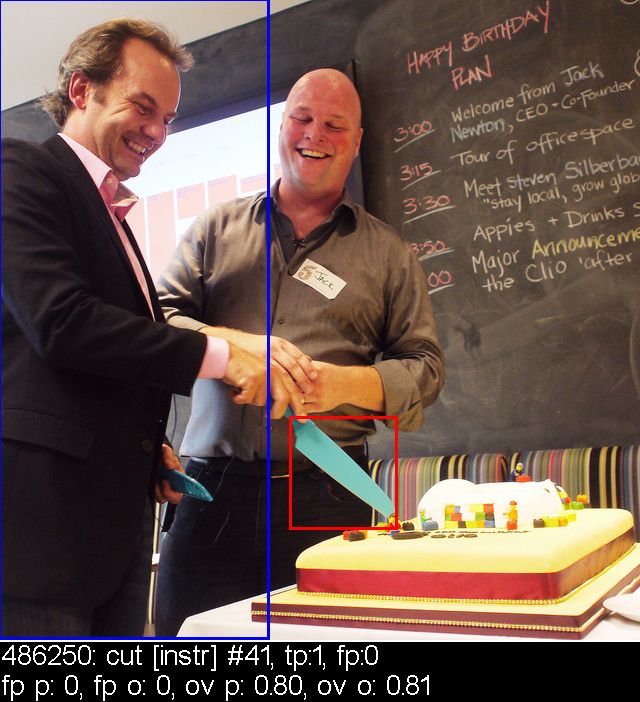}
\insertB{0.13}{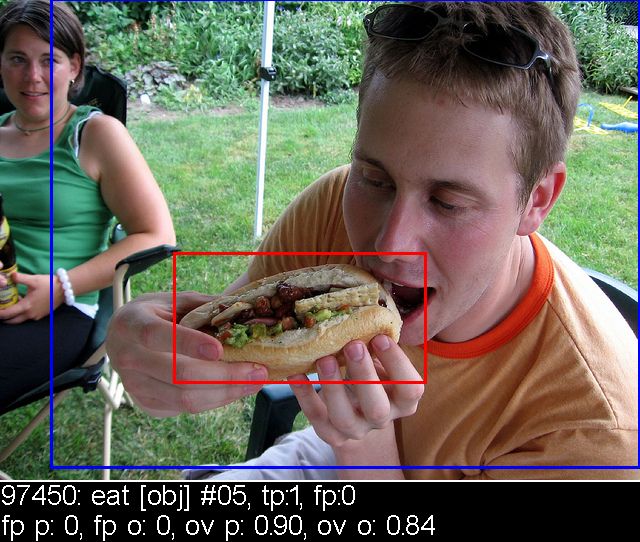}
\insertB{0.13}{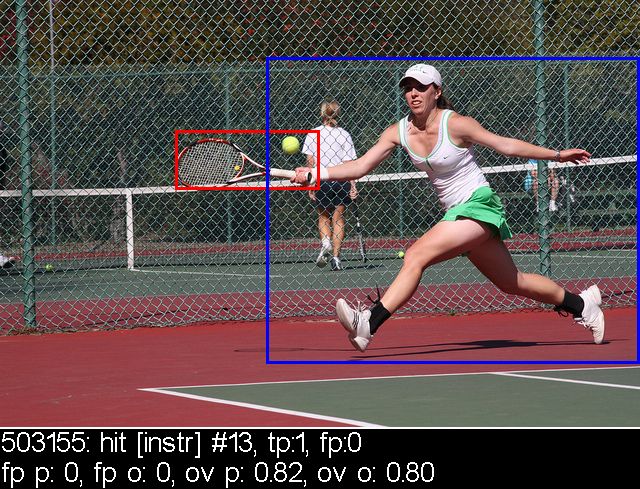}
\insertB{0.13}{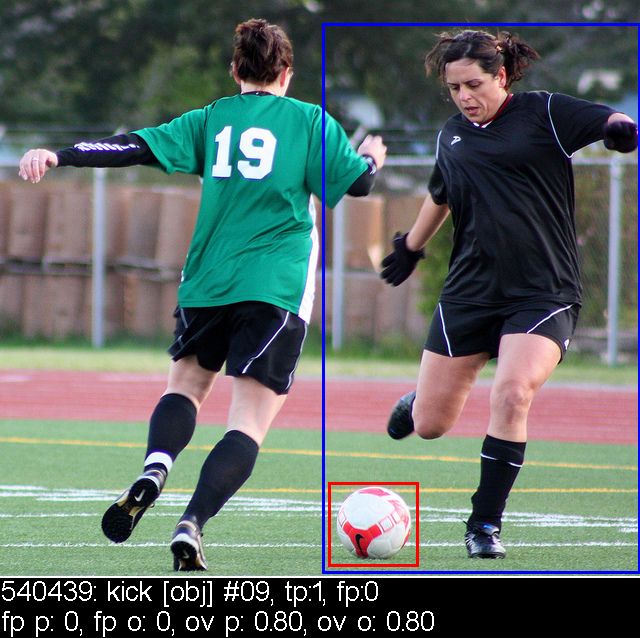}
\insertB{0.13}{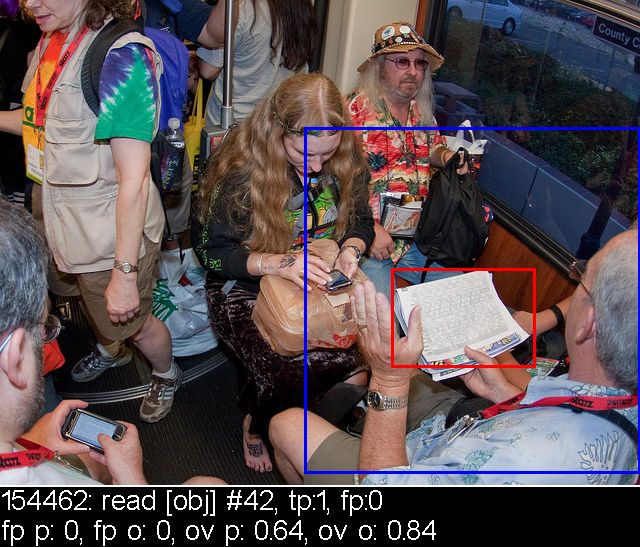}
\insertB{0.13}{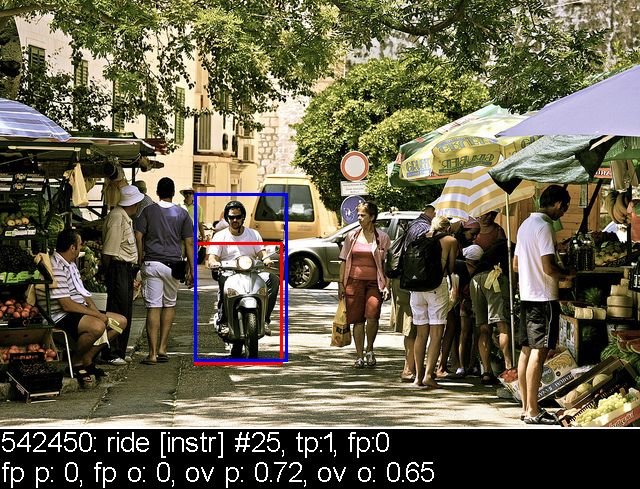}
\insertB{0.13}{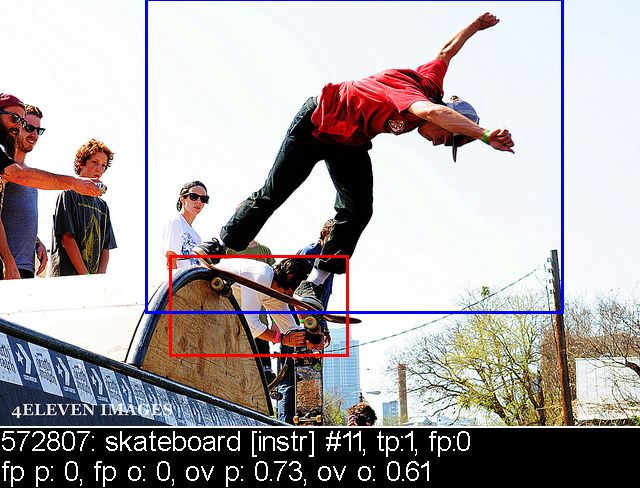}
\insertB{0.13}{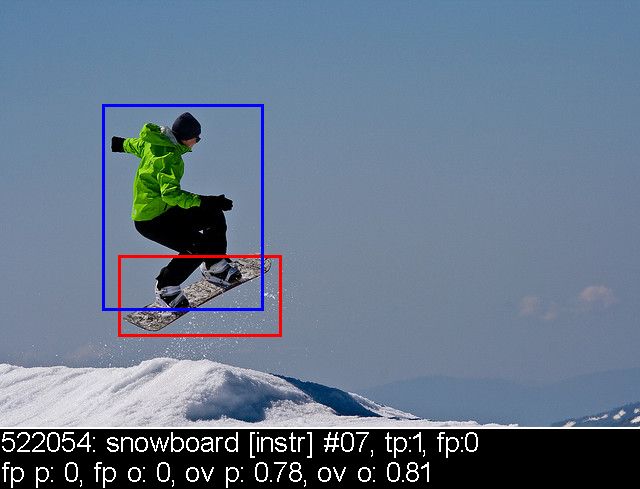}
\insertB{0.13}{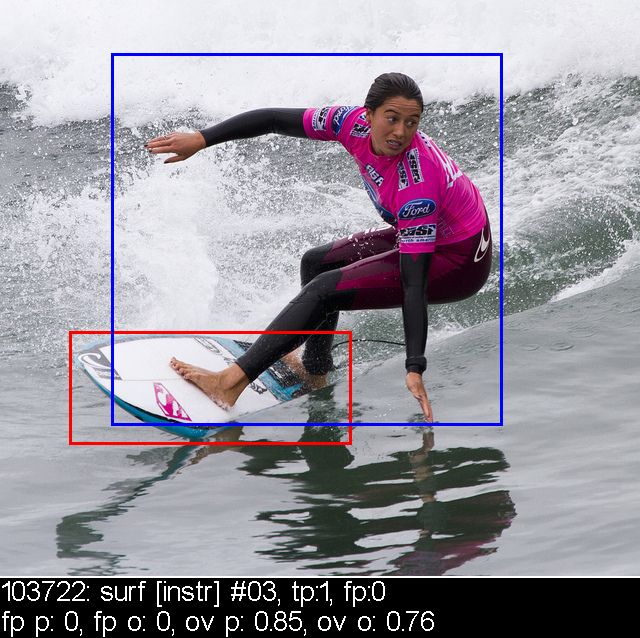}
\insertB{0.13}{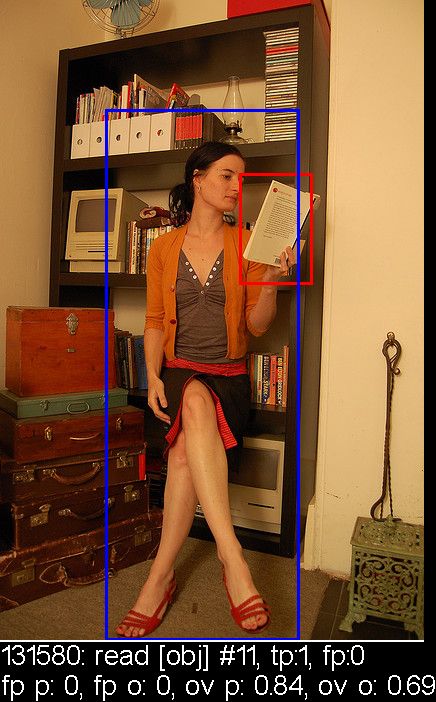}
\insertB{0.13}{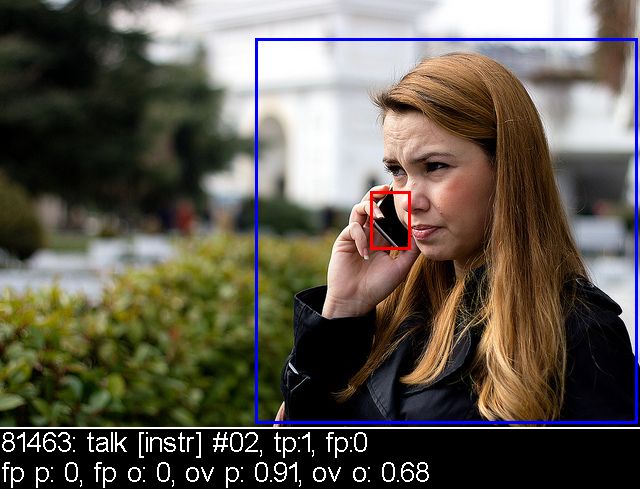}
\insertB{0.13}{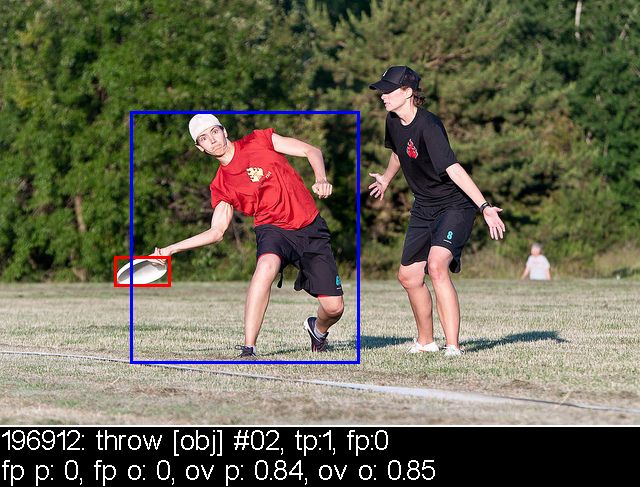}
\\ \vertical{Incorrect Class} &
\insertB{0.13}{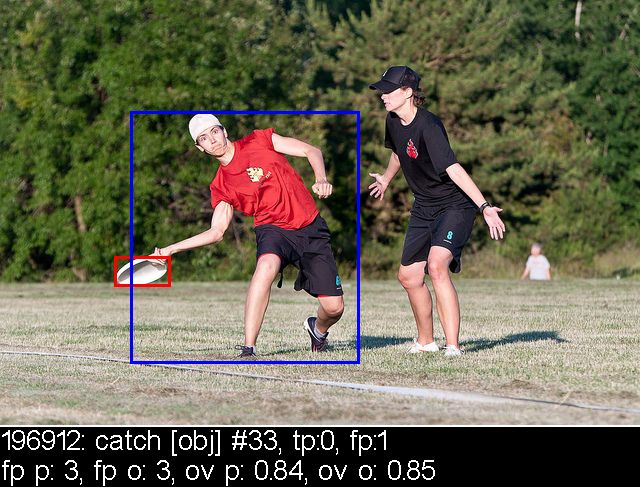}
\insertB{0.13}{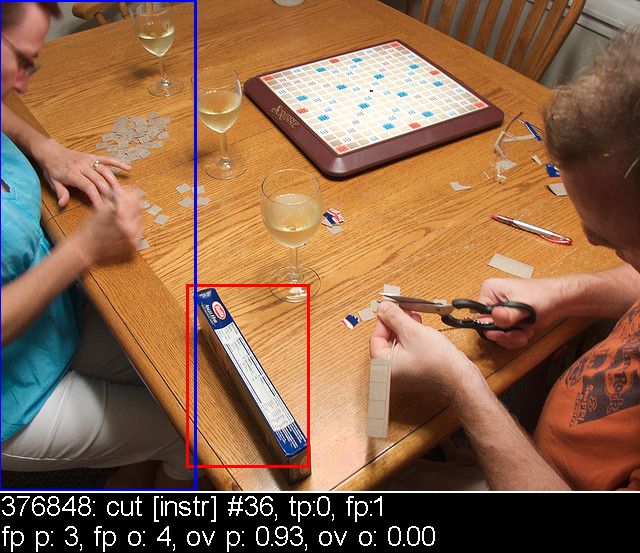}
\insertB{0.13}{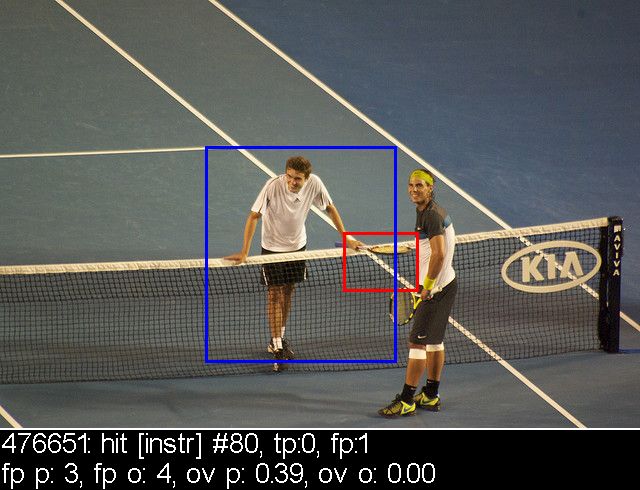}
\insertB{0.13}{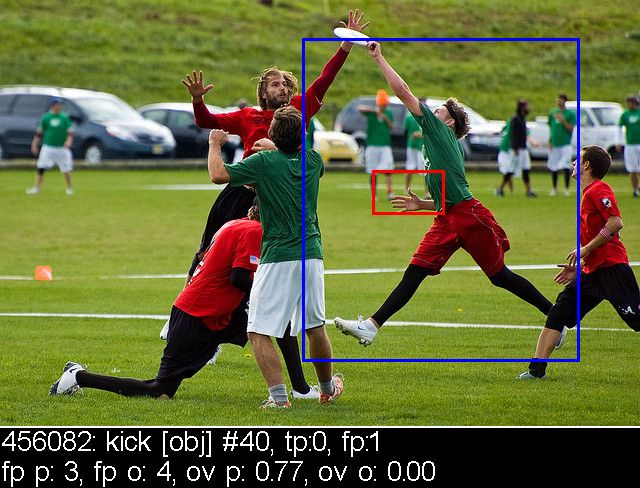}
\insertB{0.13}{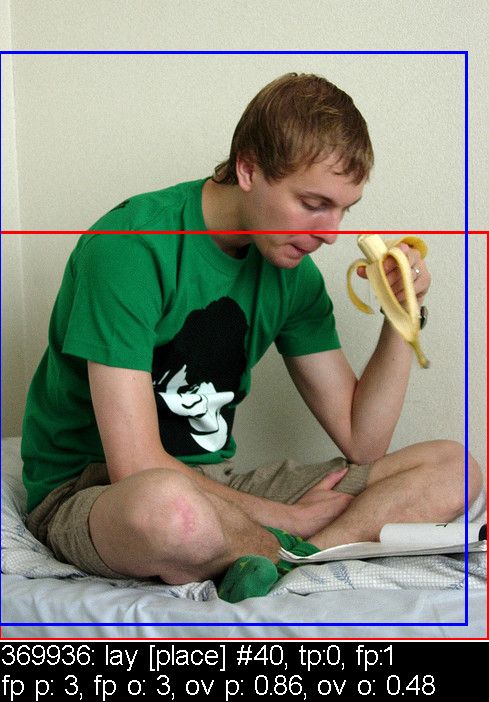}
\insertB{0.13}{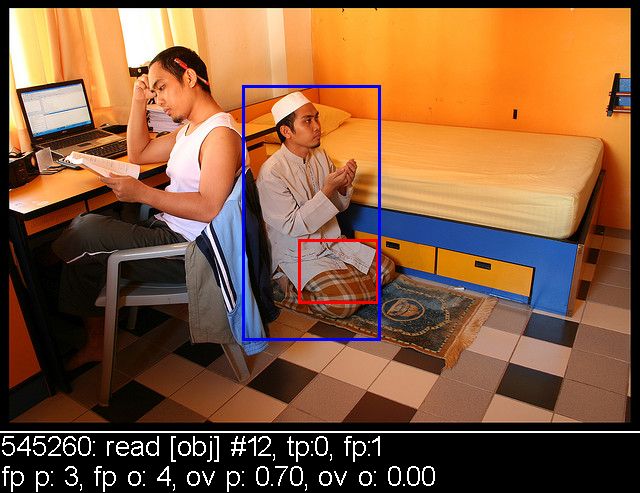}
\insertB{0.13}{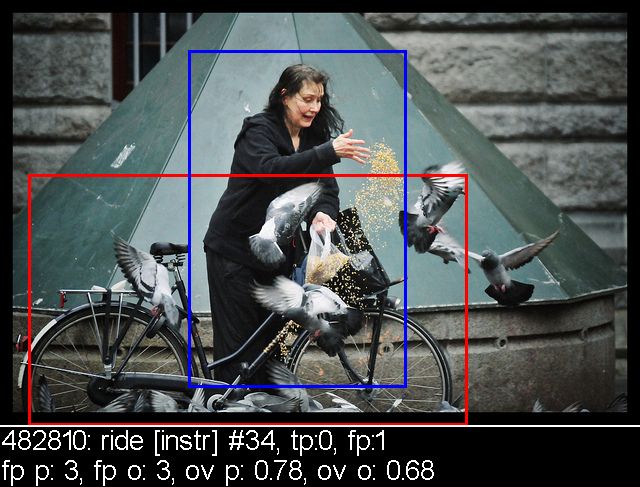}
\insertB{0.13}{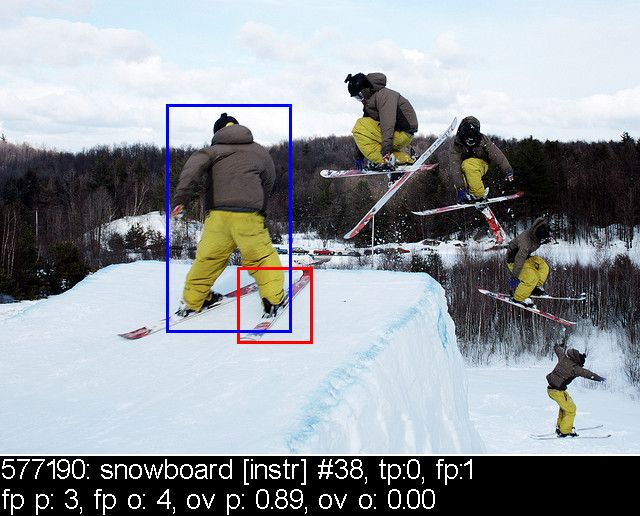}
\insertB{0.13}{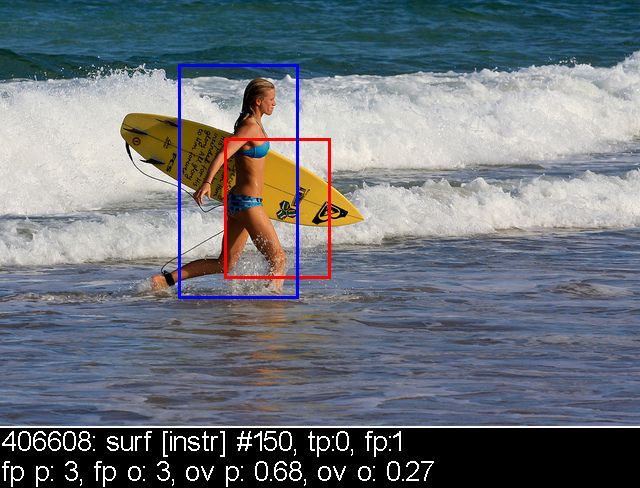}
\insertB{0.13}{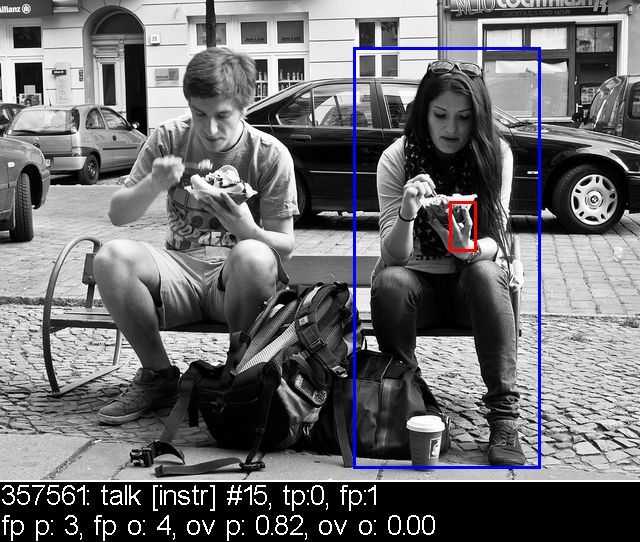}
\insertB{0.13}{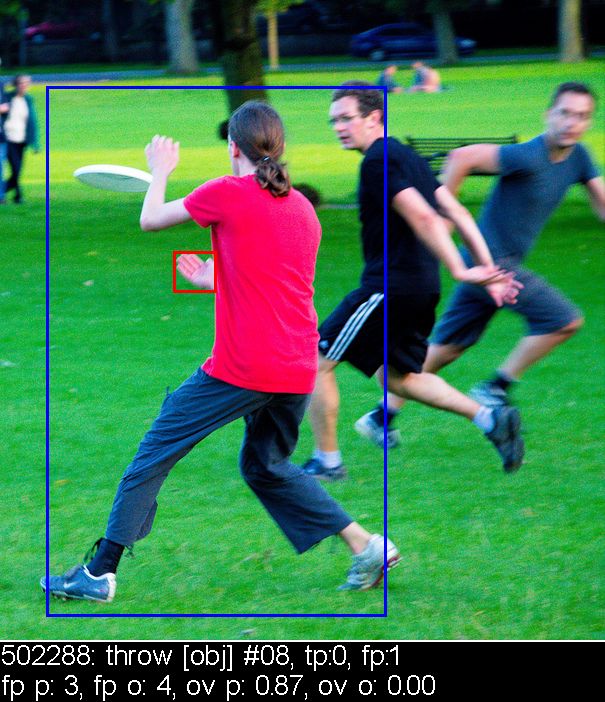}
\insertB{0.13}{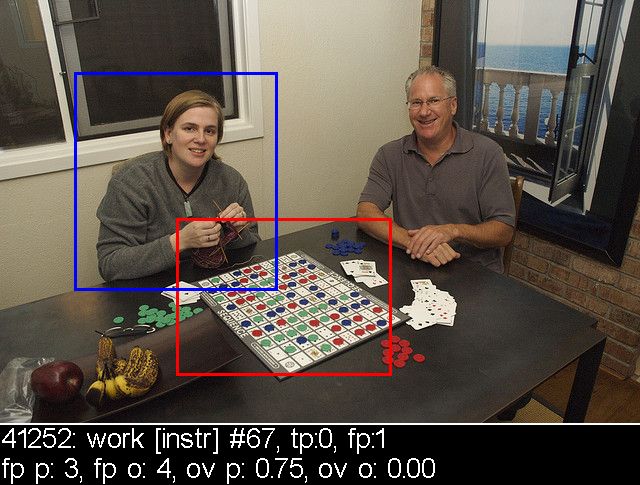}
\\ \vertical{Mis-Grouping} &
\insertB{0.13}{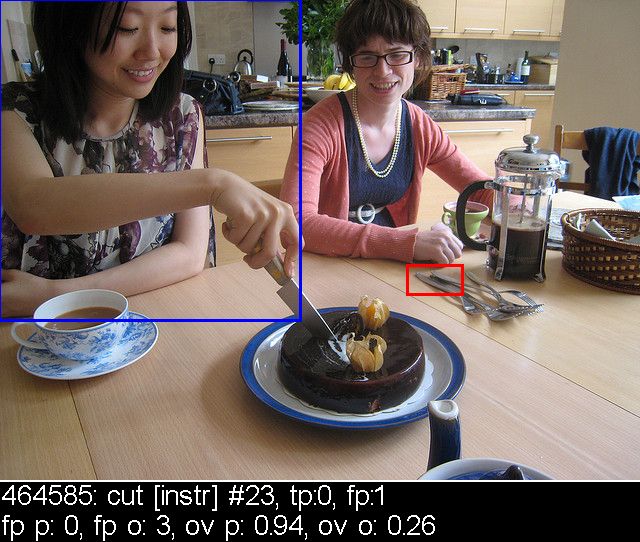}
\insertB{0.13}{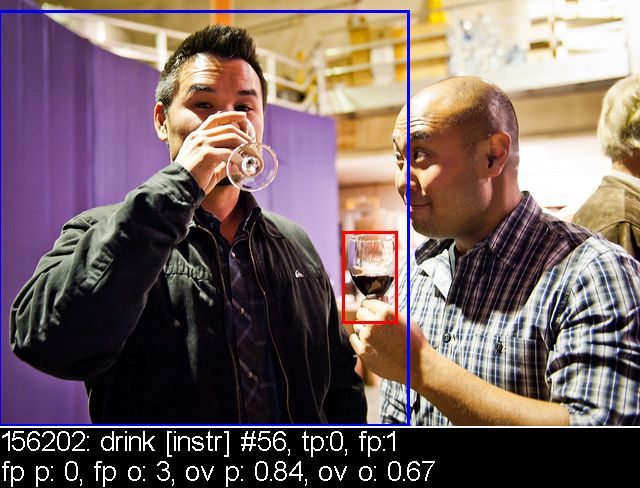}
\insertB{0.13}{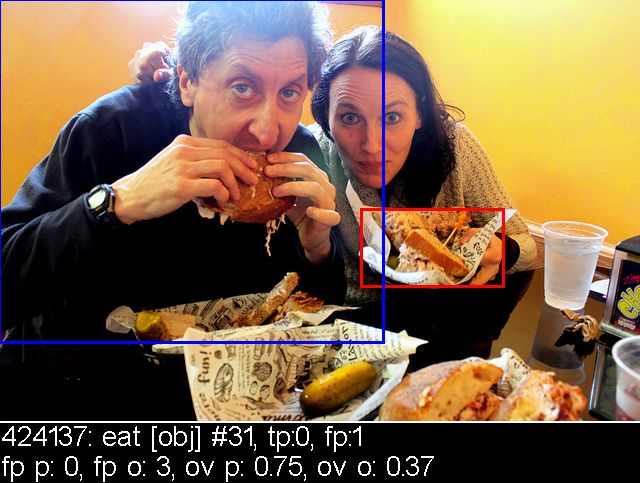}
\insertB{0.13}{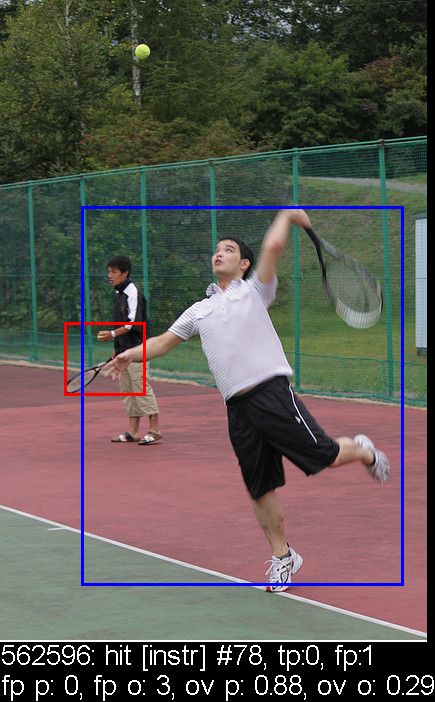}
\insertB{0.13}{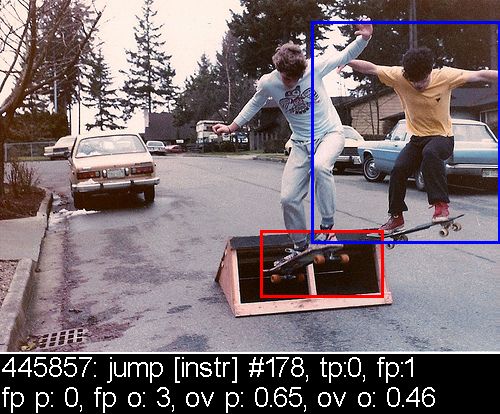}
\insertB{0.13}{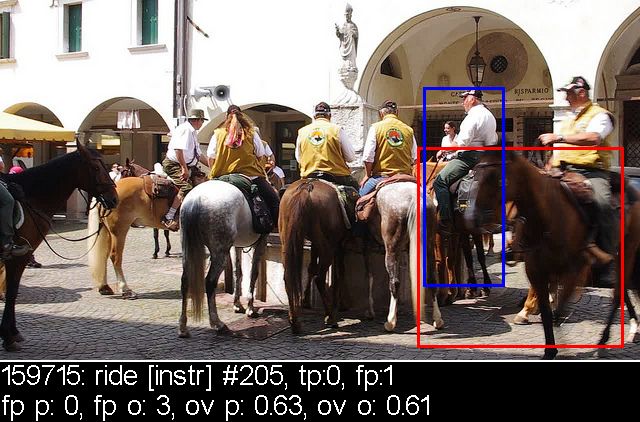}
\insertB{0.13}{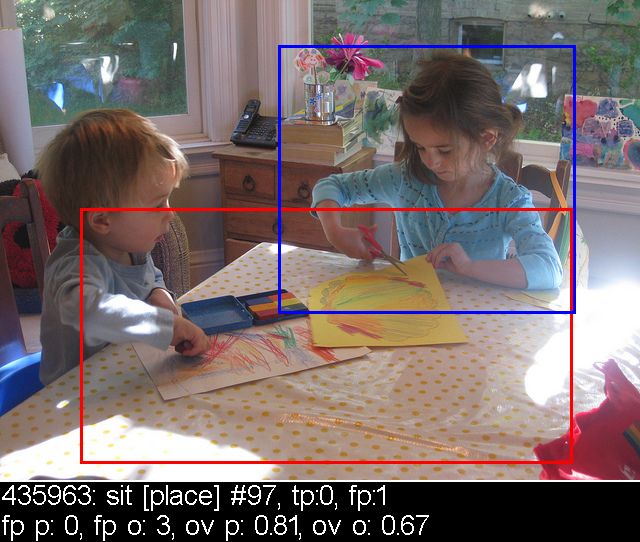}
\insertB{0.13}{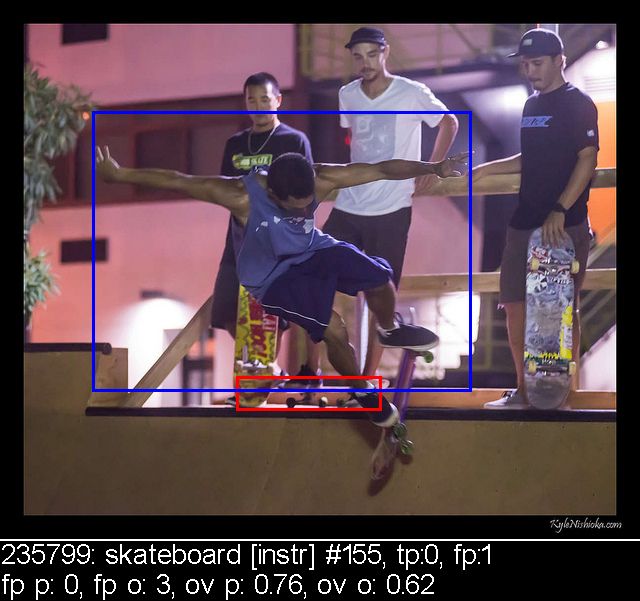}
\insertB{0.13}{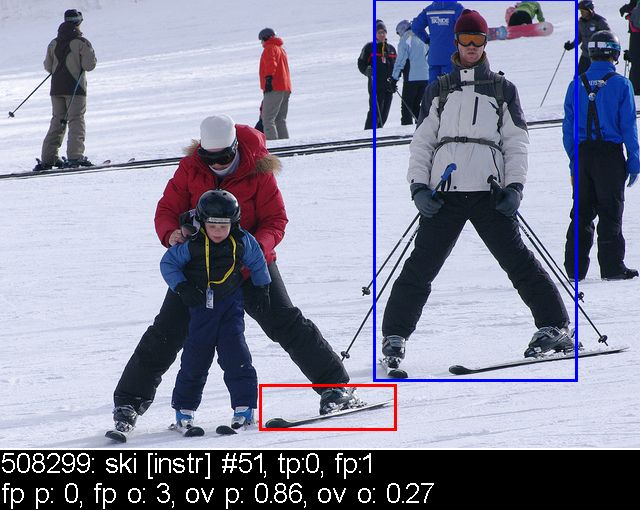}
\insertB{0.13}{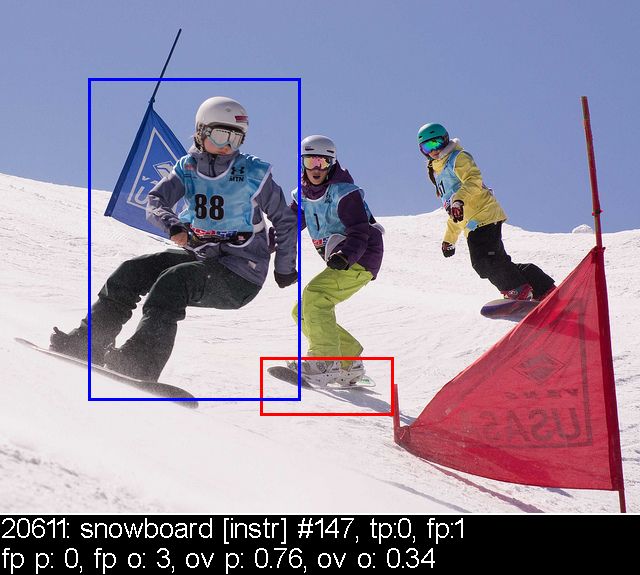}
\insertB{0.13}{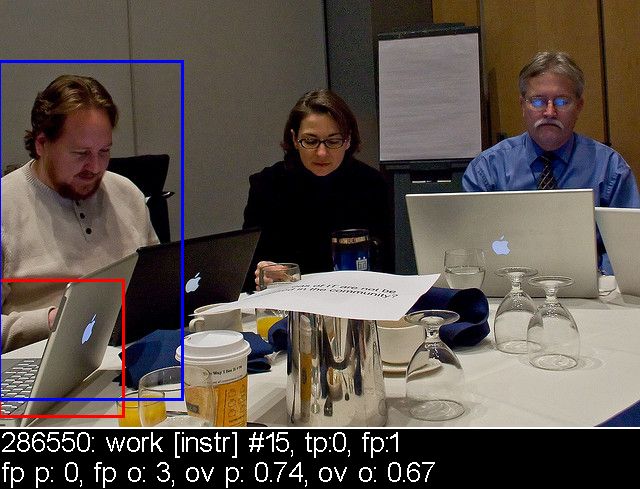}
\insertB{0.13}{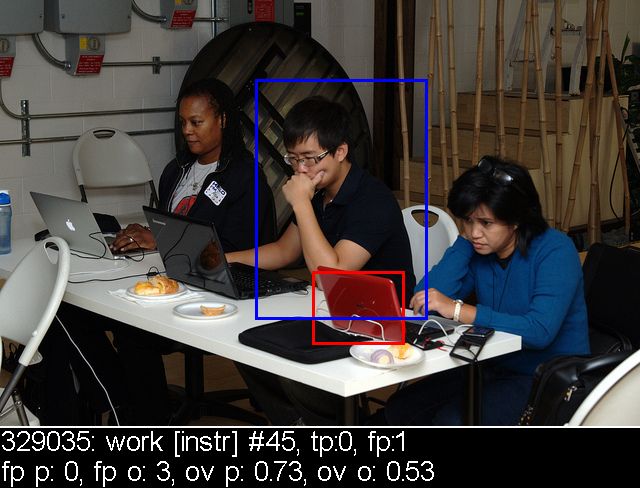}
\\ \vertical{Mis-Localization} &
\insertB{0.13}{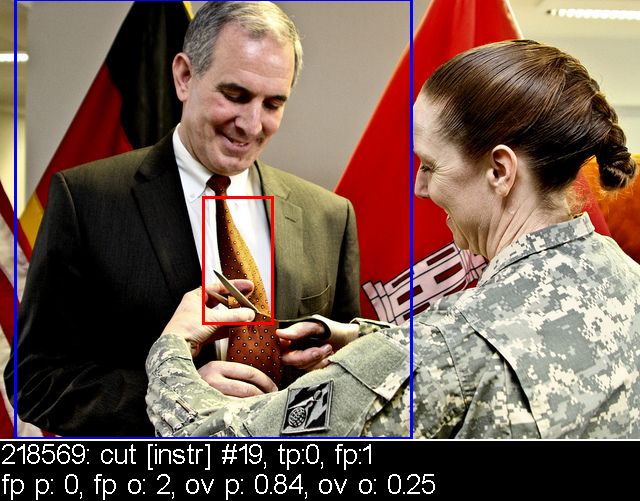}
\insertB{0.13}{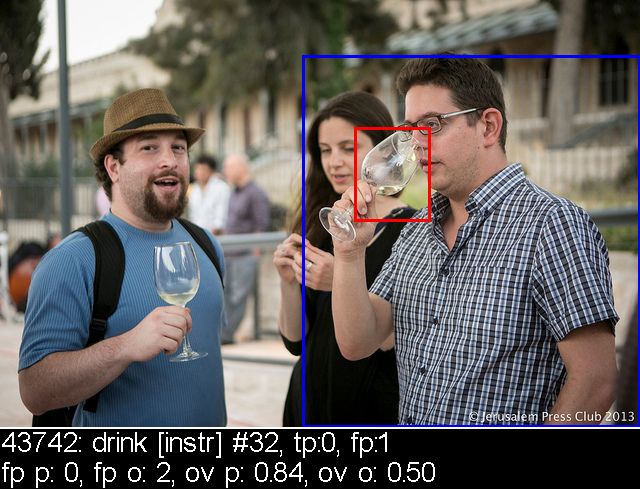}
\insertB{0.13}{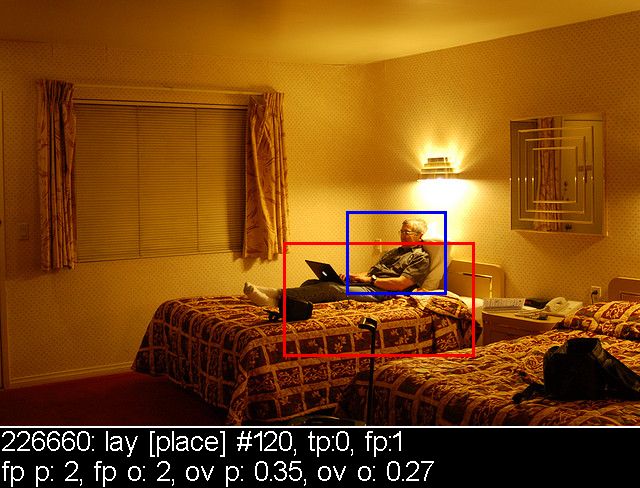}
\insertB{0.13}{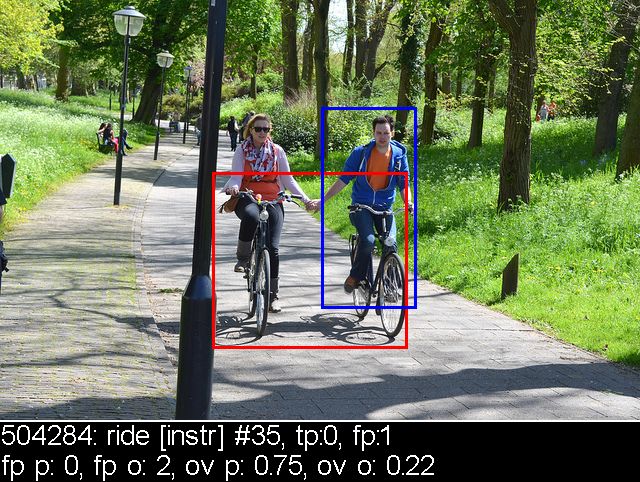}
\insertB{0.13}{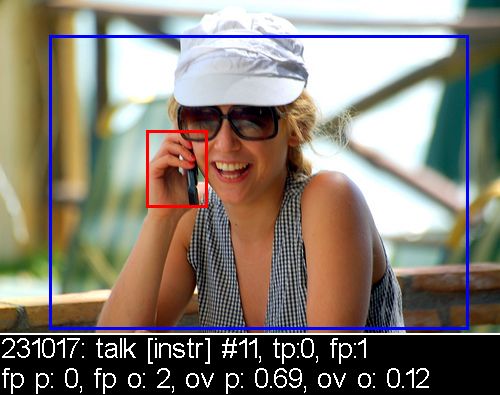}
\\ \vertical{Hallucination} &
\insertB{0.13}{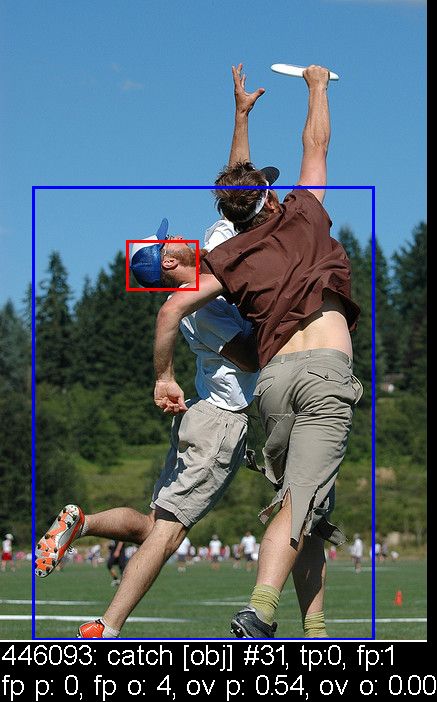}
\insertB{0.13}{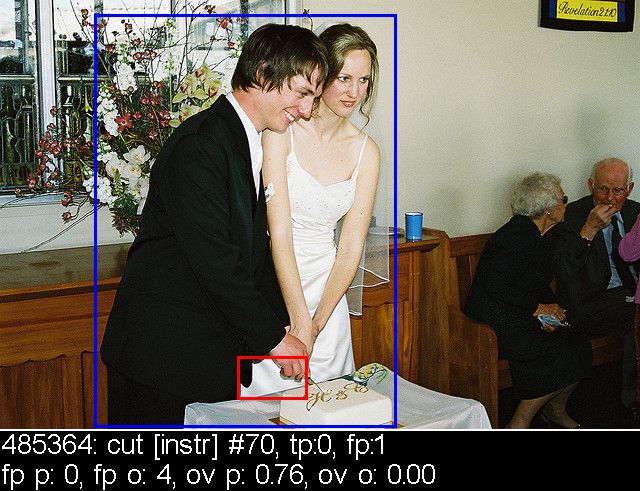}
\insertB{0.13}{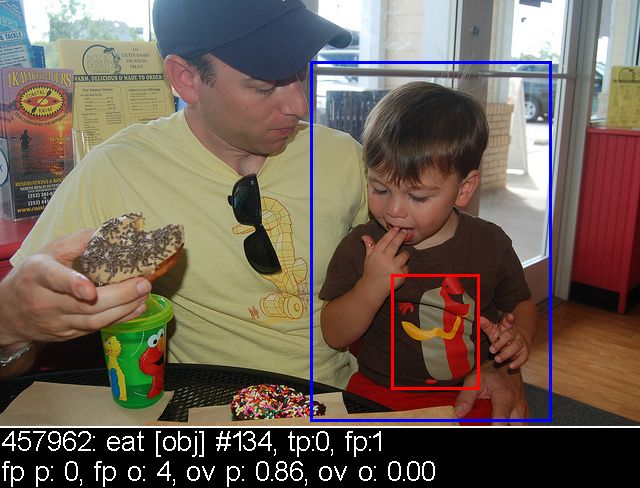}
\insertB{0.13}{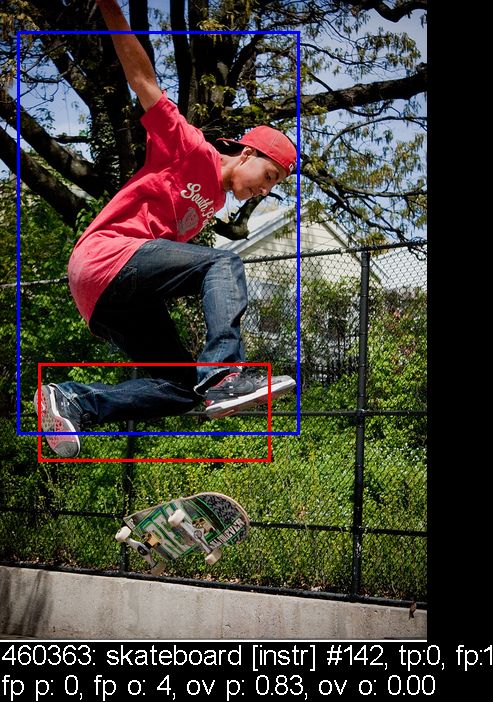}
\insertB{0.13}{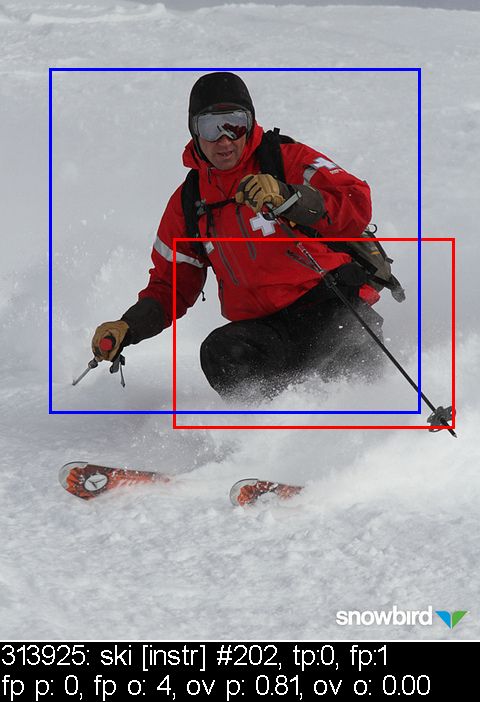}
\insertB{0.13}{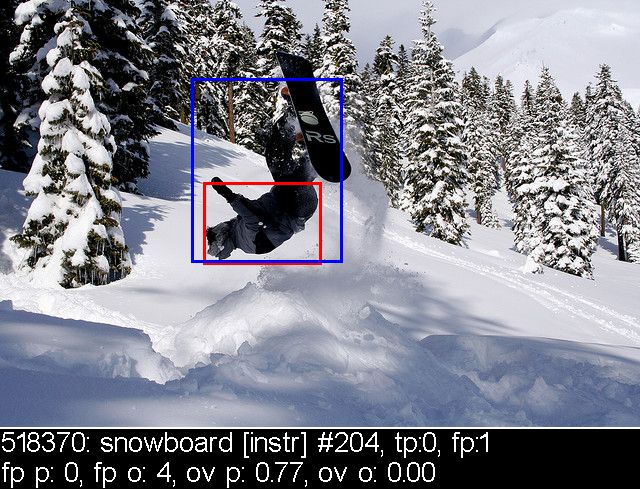}
\insertB{0.13}{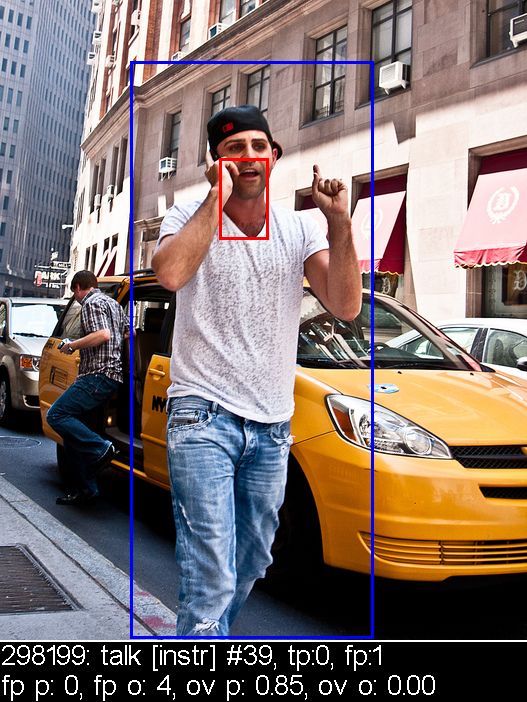}
\\
\end{tabular}
\caption{Visualizations of detections from our best performing baseline
algorithm. We show the detected agent in the blue box and the detected object
in the semantic role in the red box and indicate the inferred action class in
the test at the bottom of the image. We show some correct detections in the top
two rows, and common error modes in subsequent rows. `Incorrect Class': when
the inferred action class label is wrong; `Mis-Grouping': correctly localized
and semantically feasible but incorrectly matched to the agent; and
`Mis-localization' and `Hallucination' of the object of interaction.}
\figlabel{vis-detections}
\end{figure*}

We summarize our results here. We report all results on the \vcoco \val set.
Since we use bounding box proposals, we analyze the recall for these proposals
on the objects that are part of various semantic roles. For each semantic role
for each action class, we compute the coverage (measured as the intersection
  over union of the best overlapping MCG bounding box with the ground truth
  bounding box for the object in the role) for each instance and report the
  mean coverage, recall at 50\% overlap and recall at 70\% overlap
  (\tableref{ap} columns three to five). We see reasonable recall whenever the
  object in the semantic role is large (\eg bed, bench for \vb{lay}, horse,
  elephant, train, buses for \vb{ride}) or small but highly distinctive (\eg
  football for \vb{kick}, doughnuts, hot dogs for \vb{eat obj}) but worse when
  the object can be in drastic motion (\eg tennis rackets and baseball bats for
  \vb{hit instr}), or small and not distinctive (\eg tennis ball for \vb{hit
  obj}, cell phone for \vb{talk on phone}, ski for \vb{ski}, scissors and knife
  for \vb{cut}). 

Given that our algorithms start with a person detection, we report the
average precision of the person detector we are using. On the \vcoco \val set
our person detector which uses the 16-layer \vgg network in the Fast R-CNN
\cite{fastrcnn} framework gives an average precision of 62.54\%. 

We next report the performance at the task of agent detection (\tableref{ap})
using model $A$ as described in \secref{method}. We observe a mean average
precision of 57.5\%.  Performance is high for action classes which occur in a
distinctive scene like \vb{surf} (94.0\%, occurring in water) \vb{ski},
\vb{snowboard} (81.9\% and 75.8\%, occurring in snow) and \vb{hit} (82.6\%,
occurring in sports fields).  Performance is also high for classes which have a
distinctive object associated with the action like \vb{eat} (70.2\%).
Performance for classes which are identified by an object which is not easy to
identify is lower \eg wine glasses for \vb{drink} (25.1\%), books for \vb{read}
(10.6\%), \eg object being cut and the instrument being used for \vb{cut}
(44.5). Performance is also worse for action classes which require reasoning
about large spatial relationships \eg 61.6\% for \vb{kick}, and fine grained
reasoning of human pose \eg 41.4\%, 50.1\% for \vb{catch} and \vb{throw}. 
Finetuning the \vgg representation for this task improves performance
significantly and just training a SVM on the \vgg \texttt{fc7} features
(finetuned for object detection on \coco) performs much worse at 46.8\%.

We now report performance of the two baseline algorithms on the role detection
task. We first report performance of algorithm $B_0$ which simply pastes the box
at the mean deformation location and scale (determined using the mean of the
$\delta$ vector as defined in \eqref{delta} across the training set for each
action class separately).  This does poorly and gives a mean average precision
for the role detection task of 4.5\%. Using the regression model $B$ as
  described in \secref{method} to predict the location and scale of the
  semantic role does better giving a mAP of 7.9\%, with high performing classes
  being \vb{sit}, and \vb{lay} for which the object of interaction is always
  below the person. Using object detector output from \vgg and using the
  location of the highest scoring object detection (from the set of relevant
  categories for the semantic roles for the action class) without any spatial
  model (denoted as $C_0$) gives a mAP of 23.4\%.  Finally, model $C$ which
  also uses a spatial consistency term in addition to the score of the objects
  detected in the image performs the best among these four baseline algorithms
  giving a mAP of 26.4\%. Modeling the spatial relationship helps for cases
  when there are multiple agents in the scene doing similar things \eg
  performance for \vb{eat} goes up from 37.0\% to 46.2\%, for \vb{ride} goes up
  from 12.5\% to 27.6\%.

\paragraph{Visualizations}
Finally, we visualize the output from our best performing baseline algorithm in
\figref{vis-detections}. We show some correct detections and various error
modes. One of the common error modes is incorrect labeling of the action
(\vb{ski} \vs \vb{snowboard}, \vb{catch} \vs \vb{throw}). Even with a spatial
model, there is very often a mis grouping for the incorrect role with the
agent. This is common when there are multiple people doing the same action in
an image \eg multiple people \vb{riding} horses, or \vb{skateboarding}, or
\vb{working} on a laptops.  Finally, a lot of errors are also due to
mis-localization and hallucination of object of interaction in particular when
the object is small \eg ski for \vb{skiing}, books for \vb{reading}. 

\paragraph{Error Modes}
Having such annotations also enables us to analyze different error modes.
Following \cite{hoiem2012diagnosing}, we consider the top
$num\_inst$ detections for each class ($num\_inst$ is the number of instances
for that action class), and classify the false positives in these top
detections into the following error modes:
  \begin{enumerate}
  \item \textbf{bck}: when the agent is detected on the background. (IU with
  any labeled person is less than 0.1).
  \item \textbf{bck person}: when the agent is detected on the background, close
  to people in the background. Detections on background people are not
  penalized, however detections which have overlap between 0.1 and 0.5 with
  people in the background are still penalized and this error mode computes
  that fraction.
  \item \textbf{incorrect label}: when the agent is detected around a person
  that is labeled to be not doing this action.
  \item \textbf{person misloc}: when the agent is detected around a person
  doing the action but is not correctly localized (IU between 0.1 and 0.5) (the
  object is correctly localized).
  \item \textbf{obj misloc}: when the object in the semantic role is not
  properly localized (IU between 0.1 and 0.5) (the agent is correctly
  localized).
  \item \textbf{both misloc}: when both the object and the agent are improperly
  localized (IU for both is between 0.1 and 0.5).
  \item \textbf{mis pairing}: when the object is of the correct semantic class
  but not in the semantic role associated with this agent.
  \item \textbf{obj hallucination}: when the object is detected on the
  background.
  \end{enumerate}
\figref{fp_distr} shows the distribution of these errors for the 2 best
performing baselines that we experimented with, model $C_0$ and $C$.

The most dominant error mode for these models is incorrect classification of
the action, \figref{vis-detections} shows some examples. Another error mode is
mis localization of the object for categories like \vb{ski}, \vb{surf},
\vb{skateboard}, and \vb{snowboard}. This is also evident from the poor recall
of the region proposals for objects categories associated with these actions. A
large number of errors also come from `person misloc' for categories like
\vb{lay} which is because of unusual agent pose. We also observe that the `mis
pairing' errors decrease as we start modeling the deformation between the agent
and the object. Finally, a large number of error for \vb{cut} and \vb{hit-obj}
come from hallucinations of the object in the background.

\begin{figure*}
\centering
\begin{subfigure}{0.48\textwidth}
\insertA{1.0}{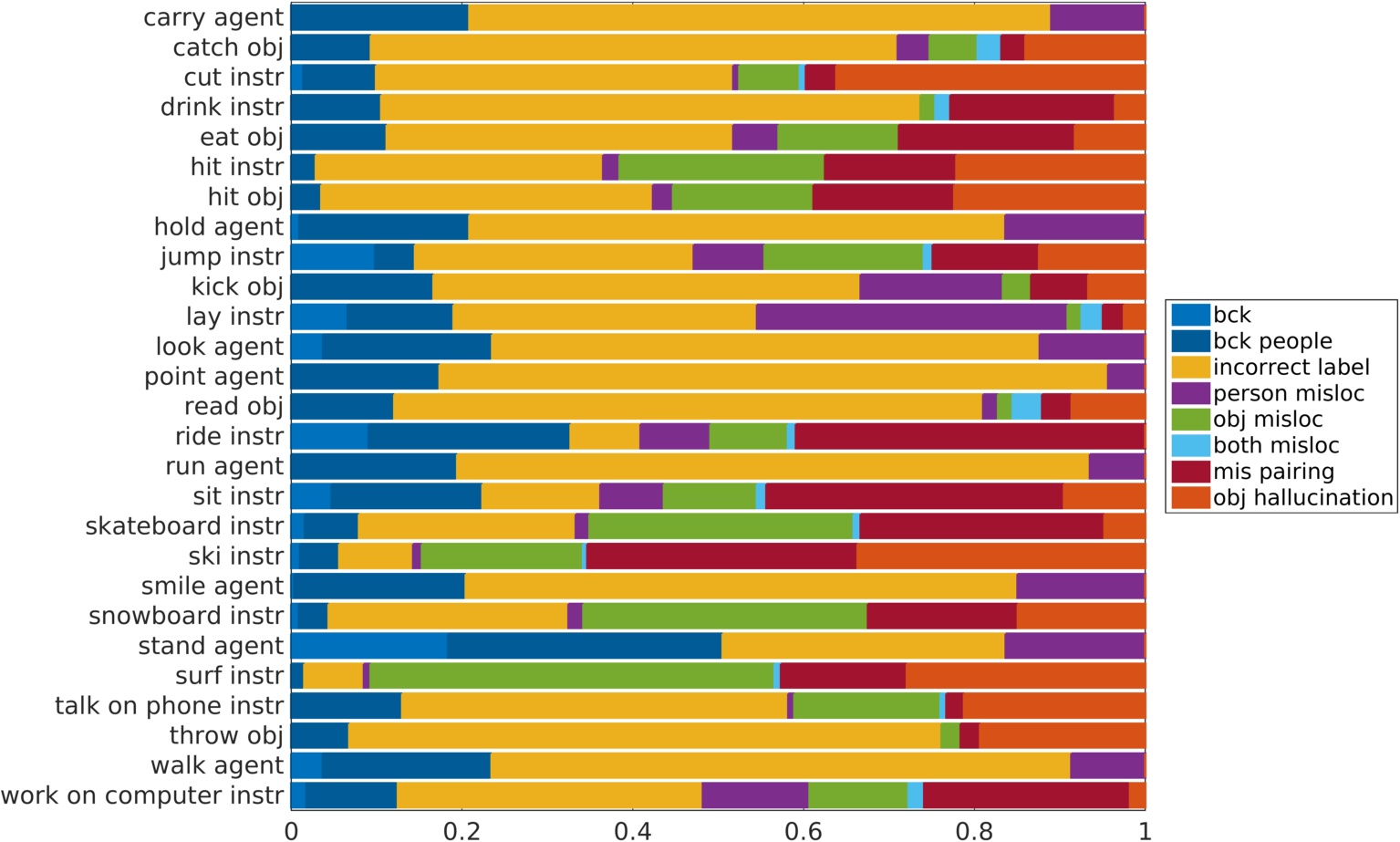}
\caption{Full Model without deformations ($C_0$)} \end{subfigure} 
\begin{subfigure}{0.48\textwidth}
\insertA{1.0}{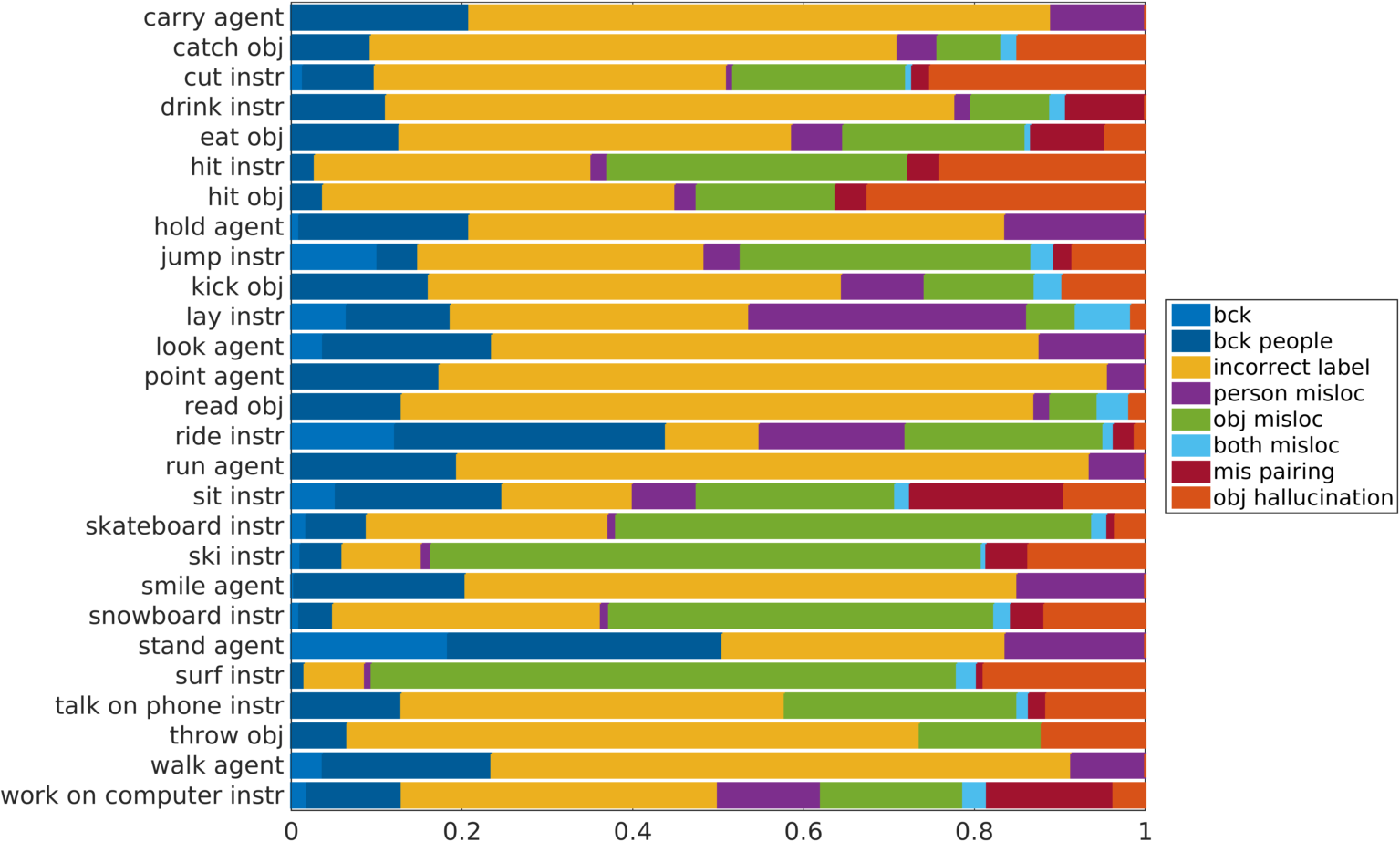}  \caption{Full
model ($C$)} \end{subfigure} 
\caption{Distribution of the false positives in the top $num\_inst$ detections
for each action class (with roles if applicable). `bck' and `bck person'
  indicate when the detected agent is on background (IU with any person less
  than 0.1) or around people in the background (IU with background people
  between 0.1 and 0.5), `incorrect label' refers to when the detected agent is
  not doing the relevant action, `person misloc' refers to when the agent
  detection is mis localized, `obj misloc' refers to when the object in the
  specific semantic role is mis localized, `both misloc' refers to when both
  the agent and the object are mis localized (mis localization means the IU is
  between 0.1 to 0.5). Finally, `mis pairing' refers to when the object of
  interaction is of the correct semantic class but not in the semantic role for
  the detected agent, and `obj hallucination' refers to when the object of
  interaction is hallucinated.  The first figure shows the distribution for the
  $C_0$ model (which does not model deformation between agent and object), 
  and the second figure shows the distribution for model $C$ (which models
  deformation between the agent and the object).}
\figlabel{fp_distr}
\end{figure*}

\paragraph{Conclusions and Future Directions}
In this work, we have proposed the task of visual semantic role labeling in
images. The goal of this task is to be able to detect people, classify what
they are doing and localize the different objects in various semantic roles
associated with the inferred action. We have collected an extensive dataset
consisting of 16K people in 10K images. Each annotated person is labeled with
26 different actions labels and has been associated with different objects in
the different semantic roles for each action. We have presented and analyzed
the performance of four simple baseline algorithms. Our analysis shows the
challenging nature of this problem and points to some natural directions of
future research. We believe our proposed dataset and tasks will enable us to
achieve a better understanding of actions and activities than current
algorithms. 

\epigraph{Concepts without percepts are empty, percepts without concepts are
blind.}{Immanuel Kant}


\paragraph{Acknowledgments: }
This work was supported by {ONR SMARTS MURI N00014-09-1-1051}, and a Berkeley
Graduate Fellowship. We gratefully acknowledge {NVIDIA} corporation for the
donation of Tesla and Titan GPUs used for this research.

{\small
\bibliographystyle{ieee}
\bibliography{refs-rbg}
}

\end{document}